\RequirePackage{fix-cm}
\documentclass[twocolumn]{svjour3}          %
\smartqed  %
\usepackage{graphicx}
\usepackage{natbib}
\usepackage[table,dvipsnames]{xcolor} %

\usepackage{multirow} %
\usepackage{booktabs} %
\usepackage{amssymb} %
\usepackage{flushend}

\usepackage{pifont}%

\usepackage{caption}
\usepackage{subcaption}

\usepackage{makecell}

\usepackage[colorlinks=true,citecolor=MidnightBlue]{hyperref}

\usepackage{makecell}

\newcommand{\updated}[1]{#1}
\newcommand{\updatedlast}[1]{#1}
\newcommand{\ingeneral}{\textbf{\updated{X-AI background.}}}
\newcommand{\indriving}{\textbf{\updated{Applications for driving models.}}}
\newcommand{\challenges}{\textbf{\updated{Challenges.}}}

\newcommand{\ie}{i.e.,\ }
\newcommand{\eg}{e.g.,\ }

\newcommand{\aka}{a.k.a.\ }

\newcommand{\lidar}{LiDAR}
\newcommand{\av}{autonomous vehicle}

\newcommand{\avs}{autonomous vehicles}
\newcommand{\Avs}{Autonomous vehicles}
\newcommand{\bev}{bird's-eye-view}

\begin{document}
\sloppy %

\title{Explainability of \updated{deep} vision-based autonomous driving systems: Review and challenges}

\author{\'Eloi Zablocki$^{*,1}$ \and 
        H\'edi Ben-Younes$^{*,1}$ \and 
        Patrick P\'erez$^{1}$ \and 
        Matthieu~Cord$^{1,2}$
        }
\authorrunning{\'Eloi Zablocki$^{*}$, H\'edi Ben-Younes$^{*}$ {\it et al.}}
        
\institute{%
\'Eloi Zablocki\\ \email{eloi.zablocki@valeo.com} \\~\\                                     
H\'edi Ben-Younes\\ \email{hedi.ben-younes@valeo.com} \\~\\
Patrick P\'erez\\ \email{patrick.perez@valeo.com} \\~\\                              
Matthieu Cord\\ \email{matthieu.cord@valeo.com} \\~\\                          
$^*$ equal contribution\\
$^1$ Valeo.ai\\                                   
$^2$ Sorbonne Université \\                                               
}

\date{}

\maketitle

\begin{abstract}

This survey reviews explainability methods for vision-based self-driving systems \updatedlast{trained with behavior cloning}. The concept of \emph{explainability} has several facets and the need for explainability is strong in driving, a safety-critical application. Gathering contributions from several research fields, namely computer vision, deep learning, autonomous driving, explainable AI (X-AI), this survey tackles several points.  First, it discusses definitions, context, and motivation for gaining more interpretability and explainability from self-driving systems, \updated{as well as the challenges that are specific to this application.}
Second, methods providing explanations to a black-box self-driving system in a post-hoc fashion are comprehensively organized and detailed. 
Third, approaches from the literature that aim at building more interpretable self-driving systems by design are presented and discussed in detail. 
Finally, remaining open-challenges and potential future research directions are identified and examined.

\end{abstract}

\rowcolors{2}{white}{gray!15}

\section{Introduction}
\label{sec:intro}

\subsection{\emph{Explainability} in the context of autonomous driving}

\subsubsection{Call for explainable autonomous driving}
\label{sec:2:xai_motivation}

The need to explain self-driving behaviors is multi-factorial. 
To begin with, autonomous driving is a high-stake and safety-critical application. It is thus natural to ask for performance guarantees, from a societal point-of-view.
However, self-driving models are not completely testable under all scenarios as it is not possible to exhaustively list and evaluate every situation the model may possibly encounter. As a fallback solution, this motivates the need for \emph{explanation} of driving decisions.

Moreover, explainability is also desirable for various reasons depending on the performance of the system to be explained. For example, as detailed by \citet{gradcam}, when the system works poorly, explanations can help engineers and researchers to improve future versions by gaining more information on corner cases, pitfalls, and potential failure modes \citep{deeptest,human_like_driving}. Moreover, when the system's performance matches human performance, explanations are needed to increase users' trust and enable the adoption of this technology \citep{lee1992trust,trusthci15,shen2020explain,trusthci20}.
In the future, if self-driving models largely outperform humans, produced explanations could be used to teach humans to better drive and to make better decisions with machine teaching \citep{teaching_categories}.

Besides, from a machine learning perspective, it is also argued that the need for explainability stems from a mismatch between training objectives on the one hand, and the more complex real-life goal on the other hand, \ie \emph{driving} \citep{mythos_interpretability,towards_rigorous_science}.
Indeed, the predictive performance on test sets does not perfectly represent performances an actual car would have when deployed to the real world.
For example, this may be due to the fact that the environment is not stationary, and the i.i.d. assumption does not hold as actions made by the model alter the environment. 
In other words, \citet{towards_rigorous_science} argue that the need for explainability arises from incompleteness in the problem formalization: machine learning objectives are flawed proxy functions towards the ultimate goal of driving. Prediction metrics alone are not sufficient to fully characterize the learned system \citep{mythos_interpretability}: extra information is needed, \emph{explanations}.
Explanations thus provide a way to check if the hand-designed objectives which are optimized enable the trained system to drive as a by-product. 

\updatedlast{Finally, as \avs{} rely more and more on deep neural networks processing visual streams \citep{cv-driving-sota}, it is of critical importance to study the explainability of driving models from a computer vision perspective. The visual input space is usually of very high dimensionality, potentially built from multiple sensor types, and it does not explicitly express any semantic concepts \citep{lime}. These considerations induce extra challenges in explaining the behavior of vision-based driving models.}

\subsubsection{Explainability: Taxonomy of terms}
\label{sec:1:xai_taxonomy}

\begin{figure}
    \includegraphics[width=\linewidth]{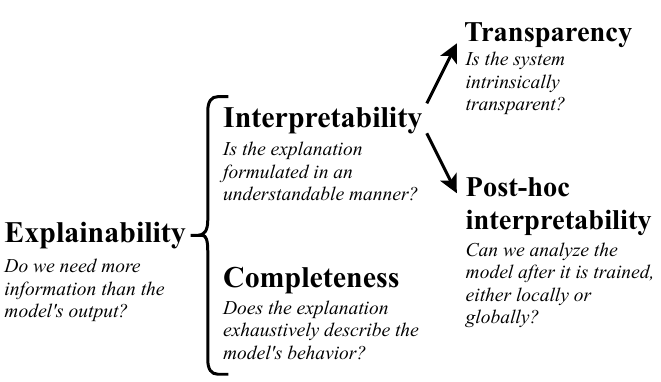}
    \caption{\textbf{Taxonomy of explainability terms adopted in this survey}. \emph{Explainability} is the combination of \emph{interpretability} (= comprehensible by humans) and \emph{completeness} (= exhaustivity of the explanation) aspects. There are two approaches to have interpretable systems: approaches intrinsic to the design of the system, which increases its \emph{transparency}, and \emph{post-hoc} approaches that justify decisions afterwards for any black-box system.}
    \label{fig:sec:2:definitions}
\end{figure}

Many terms are related to the \emph{explainability} concept and several definitions have been proposed for each of these terms. The boundaries between concepts are fuzzy and constantly evolving.
To clarify and narrow the scope of the survey, we detail here common definitions of key concepts related to explainable AI, and how they are related to one another as illustrated in \autoref{fig:sec:2:definitions}.

In human-machine interactions, \emph{explainability} is defined as the ability for the human user to understand the agent's logic \citep{xai_human_agent}. The explanation is based on how the human user understands the connections between inputs and outputs of the model. According to \citet{doshi2017accountability}, an explanation is a human-interpretable description of the process by which a decision-maker took a particular set of inputs and reached a particular conclusion. In practice, \citet{doshi2017accountability} state that an explanation should answer at least one of the three following questions: \textit{what were the main factors in the decision?} \textit{Would changing a certain factor have changed the decision?} and \textit{Why did two similar-looking cases get different decisions, or vice versa?}

The term \textit{explainability} often co-occurs with the concept of \textit{interpretability}.
While some recent work \citep{beaudouin2020identifying} advocate that the two are synonyms, \citep{GilpinBYBSK18} use the term \textit{interpretability} to designate to which extent an explanation is understandable by a human. For example, an exhaustive and completely faithful explanation is a description of the system itself and all its processing: this is a complete explanation although the exhaustive description of the processing may be incomprehensible.
\citet{GilpinBYBSK18} state that an explanation should be designed and assessed in a trade-off between its \textit{interpretability} and its \textit{completeness}, which measures how accurate the explanation is as it describes the inner workings of the system. The whole challenge in explaining neural networks is to provide explanations that are both interpretable and complete. 
\setlength\extrarowheight{5pt}

\begin{table*}[t]
    \centering
    \rowcolors{0}{white}{white}
    \begin{tabular}{@{} p{4cm} p{5cm} p{5cm} l @{}} %
        \toprule
        Who? & Why? & What? & When? \\
        \hline
        End user, citizen & Trust, situation management & Intrinsic explanations, post-hoc explanations, persuasive explanations & Before/After \\
        Designer, certification body& Debug, understand limitations and shortcomings, improve future versions, machine teaching & Stratified evaluation, corner cases, intrinsic explanations, post-hoc explanations & Before/After \\
        Justice, regulator, insurance & Liability, accountability & Exhaustive and precise explanations, complete explanations, post-hoc explanations, training and validation data & After \\
        \bottomrule 
    \end{tabular}
    \caption{\textbf{The four W's of explainable driving AI}. Who needs explanations? What kind? For what reasons? When?}
    \label{tab:2:who_what}
\end{table*}

\setlength\extrarowheight{-5pt}

\emph{Interpretability} may refer to different concepts, as explained by \citet{mythos_interpretability}. %
In particular, interpretability regroups two main concepts: model \emph{transparency} and \emph{post-hoc} interpretability. 
Increasing \emph{model transparency} amounts to gaining an understanding of \emph{how the model works}.
For example, \citet{10.1145/3236009} explain that a decision model is transparent if its decision-making process can be directly understood without any additional information; if an external tool or model is used to explain the decision-making process, the provided explanation is not transparent according to \citet{xai_human_agent}.  For \citet{trusthci15}, the system transparency can be measured as the degree to which users can understand and predict the way autonomous vehicles operate.
\updated{
On the other hand, \emph{post-hoc} interpretability relates to the fact that system comprehension is gained \emph{after} the model has been trained or after the system has been produced.
This can be the case for a specific instance, \ie local interpretability, or, more generally, to explain the whole model and/or its processing and representations.
}

An important aspect for explanations is the notion of \emph{correctness} or \emph{fidelity}. They designate whether the provided explanation accurately depicts the internal process leading to the output/decision \citep{xai_field_guide}. In the case of transparent systems, explanations are faithful by design, however, this is not guaranteed with post-hoc explanations which may be chosen and optimized their capacity to persuade users instead of accurately unveiling the system's inner workings.

Finally, it is worth mentioning that explainability in general --- and interpretability and transparency in particular --- serve and assist broader concepts such as traceability, auditability, liability, and accountability \citep{beaudouin2020identifying}.

\subsubsection{Contextual elements of an explanation}
\label{sec:2:xai_context}

The relation with \avs{} differs a lot given who is interacting with the system: 
surrounding pedestrians and end-users of the ego-car put their life in the hand of the driving system and thus need to gain trust in the system; designers of self-driving systems seek to understand limitations and shortcomings of the developed models to improve next versions; insurance companies and certification organizations need guarantees about the autonomous system.
These categories of stakeholders have varying expectations and thus the need for explanations has different motivations. The discussions of this subsection are summarized in \autoref{tab:2:who_what}.

\paragraph{Car users, citizens and trust.}

There is a long and dense line of research trying to define, characterize, evaluate, and increase the trust between an individual and a machine \citep{lee1992trust,lee1994trust,leeandsee2004,trusthci15,shariff2017psychological,du2019look,shen2020explain,trusthci20}. 
Importantly, trust is a major factor for users' acceptance of automation, as was shown in the empirical study of \citet{trusthci15}.
\citet{leeandsee2004} define trust between a human and a machine as ``\emph{the attitude that an agent will help achieve an individual's goal, in a situation characterized with uncertainty and vulnerability}''.
According to \citet{lee1992trust}, human-machine trust depends on three main factors. First, performance-based trust is built relatively to how well the system performs at its task. Second, process-based trust is a function of how well the human understands the methods used by the system to complete its task. Finally, purpose-based trust reflects the designer's intention in creating the system.

In the more specific case of autonomous driving, \citet{trusthci15} define three dimensions for trust in an \av{}. The first one is \emph{system transparency}, which refers to which extent the individual can predict and understand the operating of the vehicle. The second one is \emph{technical competence}, \ie the perception by the human of the vehicle's performance. The third dimension is \emph{situation management}, which is the belief that the user can take control whenever desired. As a consequence of these three dimensions of trust, \citet{trusthci20} propose several key factors to positively influence human trust in \avs{}. For example, improving the system performance is a straightforward way to gain more trust. Another possibility is to increase \emph{system transparency} by providing information that will help the user understand how the system functions. Therefore, it appears that the capacity to explain the decisions of an \av{} has a significant impact on user trust, which is crucial for broad adoption of this technology. Besides, as argued by \citet{haspiel2018explanations}, explanations are especially needed when users' expectations have been violated as a way to mitigate the damage. 

Research on human-computer interactions argues that the timing of explanations is important for trust. 
\citep{haspiel2018explanations,du2019look} conducted a user study showing that, to promote trust in the \av{}, explanations should be provided \emph{before} the vehicle takes action rather than after. 
Apart from the moment when the explanation should appear, \citet{xai_human_agent} advocate that users are not expected to spend a lot of time processing the explanation, which is why it should be concise and direct.
This is in line with other findings of \citet{shariff2017psychological,koo2015did} who show that although transparency can improve trust, providing too much information to the human end-user may cause anxiety by overwhelming the passenger and thus decrease trust. \updated{Lastly, \citet{towards_trustworthy_automation} show that expressing uncertainty about the autonomous system may lead to a drop of trust from users.}

\paragraph{System designers, certification, debugging and improvement of models.}

    Driving is a high-stake critical application, with strong safety requirements. 
    The concept of Operational Design Domain (ODD) is often used by carmakers to designate the conditions under which the car is expected to behave safely.
    Thus, whenever a machine learning model is built to address the task of driving, it is crucial to know and understand its failure modes, \ie in the case of accidents \citep{anticipating_accidents_dashcam,agent_centric_risk_assessment,SuzukiKAS18,KimLHS19,trafficaccident,accident_explanation_ability}, and to verify that these situations do not overlap with the ODD. 
    To this end, explanations can provide technical information about the current limitations and shortcomings of a model.
    
    The first step is to characterize the performance of the model. 
    While performance is often measured as an averaged metric on a test set, it may not be enough to reflect the strengths and weaknesses of the system. 
    A common practice is to stratify the evaluation into situations, so that failure modes could be highlighted. This type of method is used by the European New Car Assessment Program (Euro NCAP) to test and assess assisted driving functionalities in new vehicles.
    Such evaluation method can also be used at the development step, as in \citep{chauffeurnet} where authors build a real-world driving simulator to evaluate their system on controlled scenarios.
    When these failure modes are found in the behavior of the system, the designers of the model can augment the training set with these situations and re-train the model \citep{deepxplore}. 
    
    However, even if these global performance-based explanations are helpful to improve the model's performance, this virtuous circle may stagnate and not be sufficient to solve some types of mistakes.
    It is thus necessary to delve deeper into the inner workings of the model and to understand \emph{why} it makes those errors. 
    Practitioners will look for explanations that provide insights into the network's processing. Researchers may be interested in the regions of the image that were the most useful for the model's decision \citep{visualbackprop}, the number of activated neurons for a given input \citep{deeptest}, the measure of bias in the training data \citep{unbiased_look_dataset_bias}, \textit{etc}.
    
    This being said, conducting a rigorous validation of a machine learning-based system is a hard problem, mainly because it is not trivial to specify the requirements a neural network should meet \citep{borg2019safely}.

\paragraph{Regulators and legal considerations.}
   
   In the European General Data Protection Regulation (GDPR)\footnote{\url{https://eur-lex.europa.eu/legal-content/EN/TXT/HTML/?uri=CELEX:32016R0679&from=EN}}, it is stated that \updated{``data subjects''} have the right to obtain explanations from automated decision-making systems \updated{that may significantly affect them}. These explanations should provide \emph{``meaningful information about the logic involved''} in the decision-making process. Algorithms are expected to be available for the scrutiny of their inner workings (possibly through counterfactual interventions \citep{counterfactual_explanation_without_opening,shapgdpr}), and their decisions should be available for contesting and contradiction. This should prevent unfair and/or unethical behaviors of algorithms. 
   \updated{In the context of autonomous driving, the issues raised by personal data are generally related to storage and anonymization of raw data that contain people or license plates. This has indeed triggered interesting research problems, such as automatic anonymization \citep{obfuscation2018,ciagan2020} or training with anonymized data \citep{privacy_action_detection,obfusc_effects}, but they are beyond the scope of our survey.}

   Legal institutions are interested in explanations for \emph{liability} and \emph{accountability} purposes, especially when a self-driving system is involved in a car accident. 
   As noted in \citep{beaudouin2020identifying}, detailed explanations of all aspects of the decision process could be required to identify the reasons for a malfunction.
   This aligns with the guidelines towards algorithmic transparency and accountability published by the Association for Computing Machinery (ACM), which state that system auditability requires logging and record keeping \citep{simson2017acm}.
   In contrast with this \emph{local} form of explanations, a more \emph{global} explanation of the system's functioning could be required in a lawsuit. It consists in full or partial disclosure of source codes, training or validation data, or thorough performance analysis.
   It may also be important to provide information about the system's general logic that could be understandable, such as the goals of the loss function. 

    Notably, explanations generated for legal or regulatory institutions are likely to be different from those addressed to the end-user. 
    Here, explanations are expected to be exhaustive and precise, as the goal is to take a deep delve into the inner workings of the system. 
    These explanations are directed towards experts who will likely spend large amounts of time studying the system \citep{xai_human_agent}, and who are thus inclined to receive rich explanations with great amounts of detail.

\subsection{Autonomous driving: learning-based self-driving models}
\label{sec:1:driving}

\updatedlast{The development of autonomous vehicles has the potential to reduce crashes, fuel consumption, congestions, and to increase personal mobility \citep{anderson2014autonomous}.}
Research on autonomous vehicles is blooming thanks to recent advances in deep learning and computer vision \citep{imagenet_classification,deeplearning_nature}, as well as the development of autonomous driving datasets and simulators \citep{kitti,carla,CaesarBLVLXKPBB20,YuCWXCLMD20}.
The number of academic publications on this subject is rising in most machine learning, computer vision, robotics and transportation conferences, and journals.
\updatedlast{On the industry side, manufacturers are already producing cars equipped with advanced computer vision technologies for automatic lane following, assisted parking, or collision detection among other things; several automotive companies are designing prototypes with level 4 and 5 autonomy.}

\subsubsection{From modular pipelines to end-to-end learning}
\label{subsec:1:modular}

The history of autonomous driving systems started in the late '80s and early '90s with the European Eureka project called Prometheus \citep{development_machine_vision}.
This has later been followed by driving challenges proposed by the Defense Advanced Research Projects Agency (DARPA). In 2005, STANLEY \citep{thrun2006stanley} is the first autonomous vehicle to complete a Grand Challenge, which consists in a race of 142 miles in a desert area. 
Two years later, DARPA held the Urban Challenge, where autonomous vehicles had to drive in an urban environment, taking into account other vehicles and obeying traffic rules. 
BOSS won the challenge \citep{urmson2008autonomous}, driving 97 km in an urban area, with a speed up to 48 km/h. 
STANLEY, BOSS and the vast majority of the other approaches at this time \citep{perception_driven_autonomous} are systems composed of several sub-modules, each completing a very specific task. 
Broadly speaking, these sub-tasks deal with sensing the environment, forecasting future events, planning, taking high-level decisions, and controlling the vehicle.

\begin{figure*}
\includegraphics[width=\linewidth]{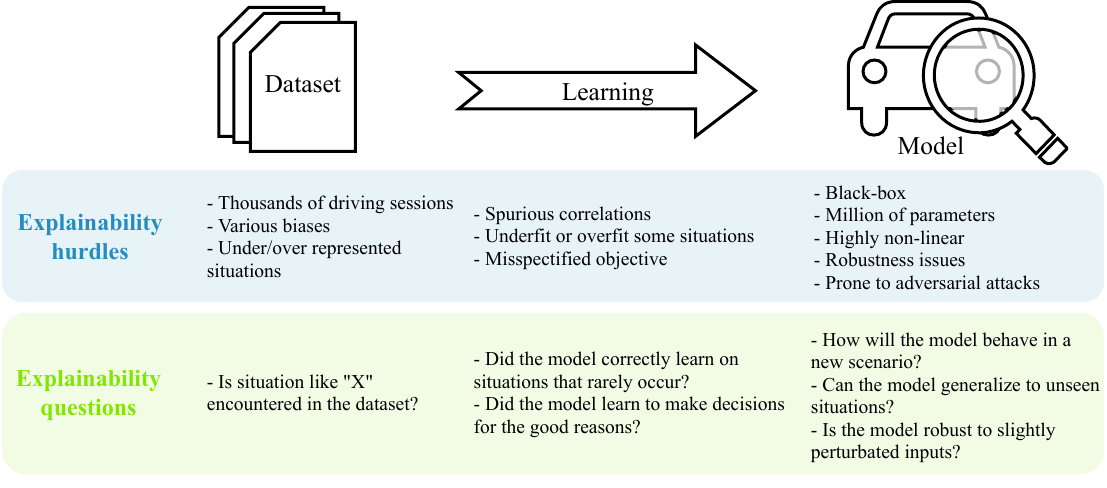}
\caption{\textbf{Explainability hurdles and questions for autonomous driving models, as seen from a machine learning point of view}.}
\label{fig:3:ml_challenges}
\end{figure*}

As pipeline architectures split the driving task into easier-to-solve problems, they offer somewhat interpretable processing of sensor data through specialized modules (perception, planning, decision, control).
However, these approaches have several drawbacks. 
First, they rely on human heuristics and manually-chosen intermediate representations, which are not proven to be optimal for the driving task. 
\updated{Second, their handcrafted nature limits their ability to account for real-world uncertainties, and therefore to generalize to scenarios that were not envisioned by system designers.}
Moreover, from an engineering point of view, these systems are hard to scale and to maintain as the various modules are entangled together \citep{latent_deep_rl}.
Finally, they are prone to error propagation between the multiple sub-modules \citep{DBLP:conf/ijcai/McAllisterGKWSC17}.

To circumvent these issues, and nurtured by the deep learning revolution \citep{imagenet_classification,deeplearning_nature}, researchers focus more and more on learning-based driving systems, and in particular on neural networks.  \updatedlast{Inspired by the seminal work of \citet{DBLP:conf/nips/Pomerleau88} who designed the first vision-based neural driving model,} 
networks are now trained either by leveraging large quantities of expert recordings \citep{pilotnet,codevilla2018end,imitation_overview} or through simulation \citep{Espi2005TORCSTO,marinaffordance,carla}.
In both cases, these systems learn a highly complex transformation that operates over input sensor data and produces end-commands (steering angle, throttle). 
While these neural driving models overcome some of the limitations of the modular pipeline stack, they are sometimes described as \emph{black-boxes} for their critical lack of transparency and interpretability. Indeed, as they are trained within the deep learning paradigm, they fall into the known shortcomings associated with these architectures. In this paper, we focus on models learned by \textit{behavior cloning}, which leverage datasets of human driving sessions, as opposed to \textit{reinforcement learning} approaches which train models through trial-and-error simulation.

\subsubsection{\Avs{} as machine learning models}
\label{sec:3:challenges:ml}

Explainability hurdles of self-driving models are shared with most deep learning models, across many application domains. Indeed, decisions of deep systems are intrinsically hard to explain as the functions these systems represent, mapping from inputs to outputs, are not transparent. 
In particular, although it may be possible for an expert to broadly understand the structure of the model, the parameter values, which have been \emph{learned}, are yet to be explained.
 
From a machine learning perspective, there are several factors giving rise to interpretability problems for self-driving systems, as machine learning researchers do not have exhaustive control over the dataset, the trained model, and the learning phase. 
\updated{Explainability methods aim at providing cues to pass these barriers, and answer some of the questions detailed in \autoref{fig:3:ml_challenges}.}

First, the dataset used for training brings interpretability problem, %
with questions such as: \emph{Has the model encounter situations like X?} Indeed, a finite training dataset cannot exhaustively cover all possible driving situations and it will likely under- and over-represent some specific ones \citep{dataset_bias}. Moreover, datasets contain numerous biases of various nature (omitted variable bias, cause-effect bias, sampling bias), which also gives rise to explainability issues related to fairness \citep{bias_fairness_ml}.

Second, the trained model, and the mapping function it represents, is poorly understood. The model is highly non-linear and does not provide any robustness guarantee as small input changes may dramatically change the output behavior. Also, these models are known to be prone to adversarial attacks \citep{fooling_car_adversarial,adversarial_attacks_defenses_driving}. Explainability issues thus occur regarding the generalizability and robustness aspects: \emph{How will the model behave under these new scenarios?}

Third, the learning phase is not perfectly understood. Among other things, there are no guarantees that the model will settle at a minimum point that generalizes well to new situations, and that the model does not underfit on some situations and overfit on others. Also, the model may learn to ground its decisions on spurious correlations during training instead of leveraging causal signals \citep{codevilla2019exploring,causal_confusion}. We aim at finding answers to questions like \emph{Which factors \emph{caused} this decision to be taken?}

\begin{figure*}
\includegraphics[width=\linewidth]{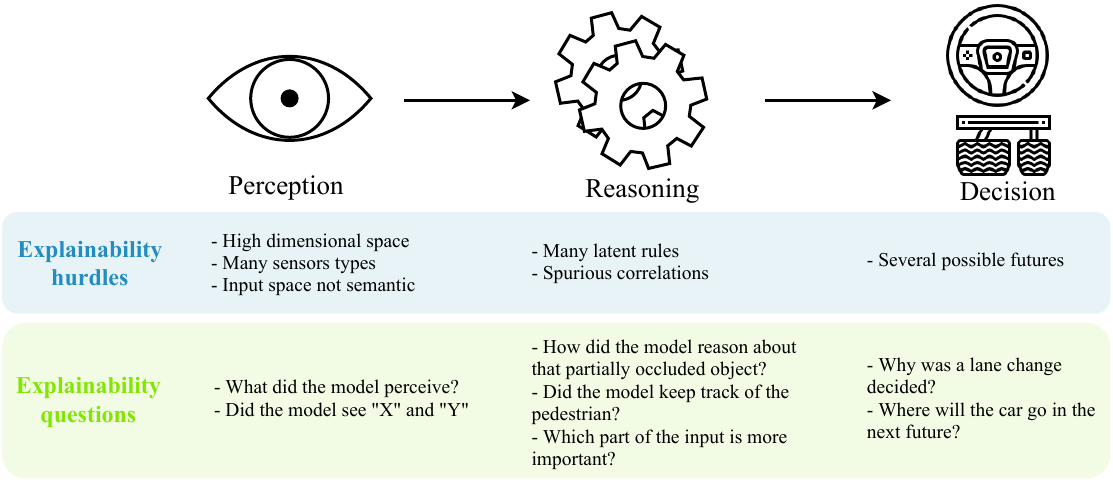}
\caption{\textbf{Explainability hurdles and questions for autonomous driving models, as seen from an autonomous driving point of view}.}
\label{fig:3:driving_challenges}
\end{figure*}

The three aforementioned barriers to model understanding are common to the vast majority learning-based systems that consume real-world data. Additionally, some specifics traits of the self-driving application should be considered. In particular, self-driving systems have to simultaneously solve intertwined tasks of very different natures, including perception tasks, motion forecasting, planning and control. Explaining a self-driving system thus means disentangling these implicit task, and to make them human-interpretable.

\subsubsection{\Avs{} as heterogeneous systems}
\label{sec:3:challenges:driving}

For humans, the complex task of driving involves solving many intermediate sub-problems, at different levels of hierarchy \citep{michon1984critical}.
In the effort towards building an autonomous driving system, researchers aim at providing the machine with these intermediate capabilities.
Thus, explaining the general behavior of \avs{} inevitably requires understanding how each of these intermediate steps is carried and how it interacts with others, as illustrated in \autoref{fig:3:driving_challenges}.
We can categorize these capabilities into three types:

\begin{itemize}
\item \textbf{Perception:} information about the system's understanding of its local environment. This includes the objects that have been recognized and assigned to a semantic label (persons, cars, urban furniture, driveable area, crosswalks, traffic lights), their localization, properties of their motion (velocity, acceleration), intentions of other agents, \textit{etc}.;
\item \textbf{Reasoning}: information about how the different components of the perceived environment are organized and assembled by the system. This includes global explanations about the rules that are learned by the model, instance-wise explanation showing which objects are relevant in a given scene \citep{visualbackprop}, traffic pattern recognition \citep{traffic_patterns}, object occlusion reasoning \citep{wojek2011occlusion,wojek2013pami};
\item \textbf{Decision:} information about how the system processes the perceived environment and its associated reasoning to produce a decision. This decision can be a high-level goal such as ``\textit{the car should turn right}'', a prediction of the ego vehicle's trajectory, its low-level relative motion or even the raw controls, \textit{etc}.
\end{itemize}

\updatedlast{While the separation between perception, reasoning, and decision is clear in modular driving systems, recent end-to-end neural networks blur the lines and perform these simultaneously \citep{pilotnet,neural_motion_planner,mp3,neat}.}
However, despite the efficiency and flexibility of end-to-end approaches, they leave small room for structured modeling of explanations, which would give the end-user a thorough understanding of how each step is achieved. 
Indeed, when an explanation method is developed for a neural driving system, it is often not clear whether it attempts to explain the perception, the reasoning, or the decision step.
Considering the nature of neural networks architecture and training, disentangling perception, reasoning, and decision in neural driving systems constitutes a non-trivial challenge.

\subsection{Survey organization}

In this survey, we organize and review \updated{deep and vision-based} self-driving models under the light of explainability. 
The scope is thus different from papers that review self-driving models in general. 
For example, \citet{geiger} review vision-based problems arising in self-driving research, \citet{di2020survey} provide a high-level review on the link between human and automated driving, \citet{imitation_overview} review imitation-based self-driving models, \citet{survey_steering_angle} survey deep learning models for predicting steering angle\updated{, \citet{taxonomy_review_driver_behavior} review approaches modeling dynamics of interactive multi-agent traffic}, and \citet{deep_rl_driving_survey} review self-driving models based on deep reinforcement learning. 

Likewise, there exist reviews on X-AI, interpretability, and explainability in machine learning in general \citep{beaudouin2020identifying,GilpinBYBSK18,xai_peeking_inside_black_box,opportunities_and_challenges_in_xai_survey}. Among others, \citet{xai_field_guide} give a pedagogic review for non-expert readers while \citet{xai_systematic_review} offer the most exhaustive and complete review on the X-AI field. \citet{xai_causal} focus on \emph{causal} interpretability in machine learning.
Moreover, there also exist reviews on explainability applied to decision-critical fields other than driving. This includes interpretable machine learning for medical applications \citep{xaimedical,xai_neuroscience}.
\updated{Concurrently to our work, the very recent survey of \citet{survey_xai_driving_omeiza} demonstrates a growing interest in the research community for explainability of autonomous driving models.}
\updated{Finally, while there are links between the fields of model \emph{validation} \citep{deeptest,adaptive_stress_testing_av,adaptive_stress_testing_reward} and model \emph{explainability}, we restrict the survey to the latter.}

Overall, the goal of this survey is diverse, and we hope that it contributes to the following:
\begin{itemize}
    \item Interpretability and explainability notions are clarified in the context of autonomous driving, depending on the type of explanations and how they are computed;
    \item Legal and regulator bodies, engineers, technical and business stakeholders can learn more about explainability methods and approach them with caution regarding presented limitations;
    \item Self-driving researchers are encouraged to explore new directions from the X-AI literature such as causality, to foster explainability and reliability of self-driving systems;
    \item The quest for interpretable models can contribute to other related topics such as fairness, privacy, and causality, by making sure that models are taking good decisions for good reasons.
\end{itemize}

Throughout the survey, we identify limitations and shortcomings from X-AI methods and propose several future research directions to have potentially more transparent, richer, and more faithful explanations for upcoming generations of self-driving models.
For the sake of simplicity and with autonomous driving research in mind, we classify methods into two main categories. 
\autoref{sec:4} presents the first category, \ie explainability methods that are applied to an already-trained deep network and are designed to provide \emph{post-hoc} explanations.
\autoref{sec:5} turns to approaches of the second category: methods providing upfront more transparency and interpretability to self-driving models' processing, by adding explainability constraints in the design of the systems. This section also presents potential future directions to increase further explainability of self-driving systems.
Finally, \autoref{sec:6} presents the particular use-case of explaining a self-driving system by means of natural language justifications.

\begin{table*}[t]
    \centering
    \rowcolors{0}{white}{white}
    \begin{tabular}{@{}l l l p{7.5cm} @{}}
        \toprule 
        Approach & Explanation type & Section & Selected references \\ %
        \hline 
        \multirow{3}{*}{Local} & Saliency map & \ref{sec:4:local:saliency} & \makecell[l]{VisualBackprop \citep{visualbackprop,explaining_pilotnet} \\ Causal filtering \citep{causal_attention} \\ Grad-CAM \citep{conditionalaffordance} \\ Meaningful Perturbations \citep{self_attention_temporal_reasoning}} \\
        & Counterfactual interventions & \ref{sec:4:local:counterfactual} & \makecell[l]{Shifting objects \citep{explaining_pilotnet} \\ Removing objects \citep{whomakedriversstop} \\ Causal factor identification \citep{chauffeurnet}} \\
        \midrule
        \multirow{1}{*}{Global} & Model translation & \ref{sec:4:global:translation} & {$\varnothing$} \\
        & Representations & \ref{sec:4:global:representations} & Neuron coverage \citep{deeptest} \\
        \bottomrule 
    \end{tabular}
    \caption{\textbf{Selected references aiming at explaining a learning-based driving model}.}
    \label{tab:explaining_model}
\end{table*}

\section{Explaining a deep driving model}
    \label{sec:4}

    When a deep learning model in general --- or a self-driving model more specifically --- comes as an opaque black-box as it has not been designed with a specific explainability constraint, \emph{post-hoc} methods have been proposed to gain interpretability from the network processing and its representations.
    Post-hoc explanations have the advantage of giving an interpretation to black-box models without conceding any predictive performance.
    In this section, we assume that we have a pre-trained model $f$. 
    Two main categories of post-hoc methods can be distinguished to explain $f$: \emph{local} methods which explain the prediction of the model for a specific instance (\autoref{sec:4:local}), and \emph{global} methods that seek to explain the model in its entirety (\autoref{sec:4:global}), \ie by gaining a finer understanding on learned representations and activations. %
    Selected references from this section are reported in \autoref{tab:explaining_model}.
    \subsection{Local explanations}
        \label{sec:4:local}
        Given an input image $x$, a \emph{local explanation} aims at justifying why the model $f$ gives its specific prediction $y=f(x)$.
        In particular, we distinguish \updated{two} types of approaches: \updated{post-hoc} saliency methods which determine regions of image $x$ influencing the most the decision (\autoref{sec:4:local:saliency})
        and counterfactual analysis which aims to find the cause in $x$ that made the model predict $f(x)$ (\autoref{sec:4:local:counterfactual}). 

        \subsubsection{\updated{Post-hoc} saliency methods}
            \label{sec:4:local:saliency}

            \ingeneral{}
            A \updated{post-hoc} \emph{saliency} method aims at explaining which input image's regions influence the most the output of the model.
            These methods produce a \emph{saliency map} (\aka \emph{heat map}) that highlights regions on which the model relied the most for its decision.
            There are \updated{three} main lines of methods to obtain a saliency map for a trained network, namely \emph{back-propagation methods}, \emph{perturbation-based methods} \updated{and \emph{local approximation} methods}.
            
            \emph{Back-propagation} methods retro-propagate output information back into the network and evaluate the gradient of the output with respect to the input, or intermediate feature-maps, to generate a heat-map of the most contributing regions. 
            These methods include DeConvNet \citep{deconvnet} and improved versions \citep{deep_inside_cnn,visual_explanation_interpretation}, Class Activation Mapping (CAM) \citep{cam}, Grad-CAM \citep{gradcam}, Layer-Wise Relevance Propagation (LRP) \citep{lrp}, deepLift \citep{deeplift}, and Integrated Gradients \citep{integrated_gradients}.
            
            \emph{Perturbation-based} methods estimate the importance of an input region by observing how modifications in this region impacts the prediction. These modifications include editing methods such as pixel \citep{deconvnet} or super-pixel \citep{lime} occlusion, greying out \citep{object_detectors_emerge_in_deep_scene} or blurring \citep{meaningful_perturbations} image regions. %
            
            \updated{\emph{Local approximation} methods approach the behavior of a trained model in the vicinity of the instance to be explained, with a simpler model. In practice, a separate model, inherently interpretable, is built to act as a proxy for the input/output mapping of the main model locally around the instance of interest. 
            In the Local Interpretable Model-agnostic Explanations (LIME) \citep{lime}, this simpler model is defined as a decision tree or a linear model, whereas if-then rules extraction is explored in \citep{ribeiro2018anchors}. SHAP \citep{shap} has been introduced to generalize LIME, and provides more consistent results.}

\begin{figure}
                \centering
                \includegraphics[width=\linewidth]{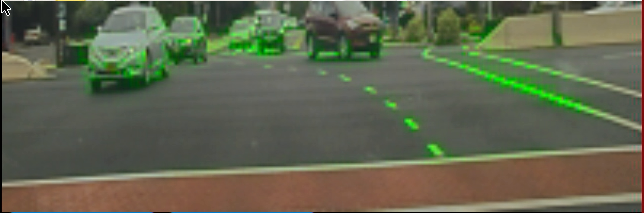}
                \caption{\textbf{Example of salient pixels \updated{(in green)} obtained by VisualBackprop \citep{visualbackprop}}.
                \updated{They highlight the central lane delimitation, the right edge of the road and other vehicles.}
                Credits to \citep{explaining_pilotnet}.
                }
                \label{fig:sec:4:visualbackprop}
            \end{figure}

           \indriving{}
            In the autonomous driving literature, \updated{post-hoc} saliency methods have been employed to highlight image regions that influence the most driving decisions. By doing so, these methods mostly explain the perception part of the driving architectures.
            The first \updated{post-hoc} saliency method to visualize the input influence in the context of autonomous driving has been developed by \citet{visualbackprop}.
            The VisualBackprop method they propose identifies sets of pixels by backpropagating activations from both late layers, which contain relevant information for the task but have a coarse resolution, and early layers which have a finer resolution. The algorithm runs in real-time and can be embedded in a self-driving car. This method has been used by \citet{explaining_pilotnet} to explain PilotNet \citep{pilotnet}, a deep end-to-end opaque self-driving architecture.
            As seen in \autoref{fig:sec:4:visualbackprop}, they qualitatively validate that the model correctly grounds its decisions on lane markings, edges of the road (delimited with grass or parked cars), and surrounding cars.  %
            
            The VisualBackprop procedure has also been employed by \citet{model_failure_saliency_maps} to gain more insights into the PilotNet architecture and its failures in particular. They use saliency maps to predict model failures by training a student model that operates over saliency maps and tries to predict the error made by the PilotNet. They find that saliency maps given by the VisualBackprop are better suited than raw input images to predict model failure, especially in case of adverse conditions.
            
            \citet{causal_attention} propose a \updated{post-hoc} saliency visualization method for self-driving models built with an attention mechanism. They explain that attention maps comprise ``blobs'' and argue that while some input blobs have a true causal influence on the output, others are spurious. Thus, they propose to segment and filter out about 60\% spurious blobs to produce simpler \emph{causal} saliency maps, derived from attention maps in a post-hoc analysis. To do so, they measure a decrease in performance when a local visual blob from an input raw image is masked out. Qualitatively, they find that the network cues on features that are also used by humans while driving, including surrounding cars and lane markings for example. %
            
            Recently, \citet{conditionalaffordance} propose to condition the \updated{post-hoc} saliency visualization on a variety of driving features, namely driving ``affordances''. They employ the Grad-CAM saliency technique \citep{gradcam} on an end-to-mid self-driving model trained to predict driving affordances on a dataset recorded from the CARLA simulator \citep{carla}. They argue that \updated{post-hoc} saliency methods are particularly well suited for this type of architecture on the contrary to end-to-end models, as all of the perception (\eg detection of speed limits, red lights, cars, \textit{etc}.) is mapped to a single control output for those models. Instead, in their case, they can analyze the \updated{post-hoc} saliency in the input image for each affordance, \eg ``hazard stop'' or ``red light''.

            Still in the context of driving scenes, although not properly for explaining a self-driving model, it is worth mentioning that \citet{self_attention_temporal_reasoning} use the perturbation-based masking strategy of \citet{meaningful_perturbations} to obtain saliency maps for a driving scene classification model trained on the HDD dataset \citep{RamanishkaCMS18}.

            \challenges{}
            While \updated{post-hoc} saliency methods enable visual explanations for deep black-box models, they come with some limitations.
            First, they are hard to evaluate. For example, human evaluation can be employed \citep{lime,crowdsourcing_evaluation_saliency} but this comes with the risk of selecting methods which are more \emph{persuasive}, \ie plausible and convincing and not necessarily \emph{faithful}.
            
            Another possibility to evaluate \updated{post-hoc} saliency methods is to use 
            \updated{automated metrics. 
            \updatedlast{In the work of \citet{SamekBMLM17}, saliency maps are evaluated by measuring the decrease of the output score induced by the removal of salient pixels of the validation set. Closely related, the Remove and Retrain (ROAR) scheme \citep{benchmark_interpretability_deep_nn} hides the most salient pixels of training images in addition to the validation set, and re-trains image classifier on this modified training set. The intuition is that the new model should perform poorly as it has been trained without salient regions that were considered important for the task. 
            Other methods involve additional ground-truth locations of image regions that cause the output. These annotations can either be annotated by humans \citep{meaningful_perturbations}, or obtained automatically with synthetic data \citep{visual_explanation_interpretation,clever-xai}.
            }}
            Second, \citet{sanity_check_saliency} indicate that the generated heat maps may be misleading as some \updated{post-hoc} saliency methods are independent both of the model and the data. Indeed, they show that some \updated{post-hoc} saliency methods behave like edge-detectors even when they are applied to a randomly initialized model.
            Third, \citet{interpretation_fragile} show that it is possible to attack visual saliency methods so that the generated heat-maps do not highlight important regions anymore, while the predicted class remains unchanged.
            Lastly, different saliency methods produce different results and it is not obvious to know which one is correct, or better than others. In that respect, a potential research direction is to learn to combine explanations coming from various explanation methods. 
            
            \updated{Interestingly, these saliency methods provide explanations that can be processed very rapidly, as they are visual in essence. 
            Thus, they constitute relevant candidates to be shown to regular car users, who will not spend too much time analyzing explanations. 
            However, as it has been discussed in \autoref{sec:2:xai_context}, the purpose of these explanations geared towards car users is mainly to build and enhance trust in the system. Considering the limitations we just discussed, a special care should be given in designing and providing saliency maps that indeed reflect the true behavior of the model.}

        \subsubsection{Counterfactual explanation}
            \label{sec:4:local:counterfactual}

            \ingeneral{}
            Recently, a lot of attention has been put on \textit{counterfactual analysis}, a field from the causal inference literature \citep{pearl_causality,xai_causal}.
            A counterfactual analysis aims at finding features $X$ within the input $x$ that \emph{caused} the decision $y=f(x)$ to be taken, by imagining a new input instance $x'$ where $X$ is changed and a different outcome $y'$ is observed. The new imaginary scenario $x'$ is called a \emph{counterfactual example} and the different output $y'$ is a contrastive class. The new counterfactual example, and the change in $X$ between $x$ and $x'$, constitute \emph{counterfactual explanations}.
            In other words, a counterfactual example is a modified version of the input, in a minimal way, that changes the prediction of the model to the predefined output $y'$. For instance, in an autonomous driving context, it corresponds to questions like ``What should be different in this scene, such that the car would have stopped instead of moving forward?''
            \updated{
            We highlight the difference between saliency methods and counterfactual explanations. Saliency methods answer a ``\emph{where} question'' as they are limited to find input regions that are the most influential for the decision. On the contrary, counterfactual explanations answer a more precise ``\emph{what} question'' as they seek minimal modifications of the input that switch the decision of the model.
            }

            Several requirements should be imposed to find counterfactual examples.
            First, the prediction $f(x')$ of the counterfactual example must be close to the desired contrastive class $y'$.
            Second, the counterfactual change must be \emph{minimal}, \ie the new counterfactual example $x'$ must be as similar as possible to $x$, either by making sparse changes or in the sense of some distance.
            Third, the counterfactual change must be \emph{relevant}, \ie new counterfactual instances must be likely in the underlying input data distribution.
            The simplest strategy to find counterfactual examples is the naive trial-and-error strategy, which finds counterfactual instances by randomly changing input features.
            More advanced protocols have been proposed, for example \citet{counterfactual_explanation_without_opening} propose to minimize both the distance between the model prediction $f(x')$ for the counterfactual $x'$ and the contrastive output $y'$ and the distance between $x$ and $x'$.

            Traditionally, counterfactual explanations have been developed for classification tasks, with a low-dimensional semantic input space, such as the credit application prediction task \citep{counterfactual_explanation_without_opening}.
            It is worth mentioning that there also exist \emph{model-based} counterfactual explanations which aim at answering questions like ``What decision would have been taken if this model component was not part of the model or designed differently?'' \citep{explaining_deep_causal_inference,causal_learning_autoencoded_activations}. 
            To tackle this task, the general idea is to model the deep network as a Functional Causal Model (FCM) on which the causal effect of a model component can be computed with causal reasoning on the FCM \citep{pearl_causality}. For example, this has been employed to gain an understanding of the latent space learned in a variational autoencoder (VAE) or a generative adversarial network (GAN) \citep{counterfactuals_generative_models}, or in RL to explain agent's behavior with counterfactual examples by modeling them with an SCM \citep{rl_causal_lens}.
            Counterfactual explanations have the advantage that they do not require access to the dataset nor the model to be computed. This aspect is important for automotive stakeholders who own datasets and industrial property of their model and who may lose a competitive advantage by being forced to disclose them. Moreover, counterfactual explanations are GDPR compliant \citep{counterfactual_explanation_without_opening}.
            A potential limit of counterfactual explanations is that they are not unique: distinct explanations can explain equally well the same situation while contradicting each other.
            
            When dealing with a high-dimensional input space --- as it is the case with images and videos --- counterfactual explanations are very challenging to obtain as naively producing examples under the requirements specified above leads to new instances $x'$ that are imperceptibly changed with respect to $x$ while having output $y'=f(x')$ dramatically different from $y=f(x)$. This can be explained given that the problem of adversarial perturbations arises with high dimensional input space of machine learning models, neural networks in particular \citep{intriguing_properties_nn}.
            To mitigate this issue in the case of image classification, \citet{counterfactual_visual_explanations} use a specific instance, called a \emph{distractor} image, from the predefined target class and identify the spatial region in the original input such that replacing them with specific regions from the distractor image would lead the system to classify the image as the target class.
            In addition, \citet{grounding_visual_explanations} provide counterfactual explanations by staying at the attribute level and by augmenting the training data with negative examples created with hand-crafted rules. %
            
            \updated{
            The automatic evaluation of counterfactual explanations must be conducted on several fronts.
            First, counterfactual changes must change the decision of the main model.
            Second, they must be realistic and the resulting altered image or video must belong to the data distribution, as measured with the Frechet Inception Distance (FID) \citep{fid}.
            Lastly, counterfactual changes must be minimal and sparse, for example, as measured by an oracle model estimating the number of attribute changes between two inputs \citep{progressive_exaggeration,dive}.
            }
            
            \begin{figure*}
                 \centering
                 \begin{subfigure}[h]{0.327\linewidth}
                     \centering
                     \captionsetup{font=footnotesize}
                     \caption{Query image}
                     \vspace{-0.2cm}
                     \includegraphics[width=\linewidth]{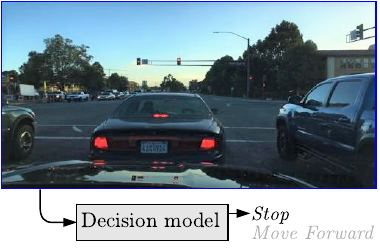}
                     \label{fig:bigpicture:a}
                 \end{subfigure}
                 \hfill
                 \begin{subfigure}[h]{0.327\linewidth}
                     \centering
                     \captionsetup{font=footnotesize}
                     \caption{Counterfactual explanation}
                     \vspace{-0.2cm}
                     \includegraphics[width=\linewidth]{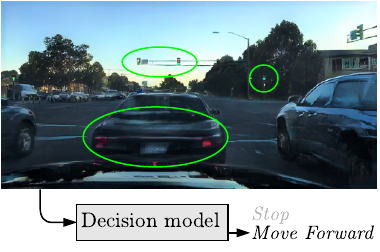}
                     \label{fig:bigpicture:b}
                 \end{subfigure}
                 \hfill
                 \begin{subfigure}[h]{0.327\linewidth}
                    \centering
                    \captionsetup{font=footnotesize}
                    \caption{\textbf{Traffic light}-targeted counterf.\ expl.}
                     \vspace{-0.2cm}
                    \includegraphics[width=\linewidth]{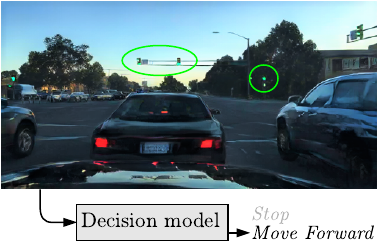}
                    \label{fig:bigpicture:d}
                 \end{subfigure}
                    \vspace{-0.5cm}
                    \caption{
                    \textbf{Counterfactual explanations generated by STEEX \citep{steex}.}
                    The `Decision model' is a binary classifier that predicts whether or not it is possible to move forward.
                    To explain that model, STEEX generates counterfactual explanations (b).
                    It can also provide explanations that target a specified region (c).
                    }
                    \label{fig:sec:4:steex}
        \end{figure*}
        
         \begin{figure}
                \centering
                \includegraphics[trim={10 25 30 10},clip,width=\linewidth]{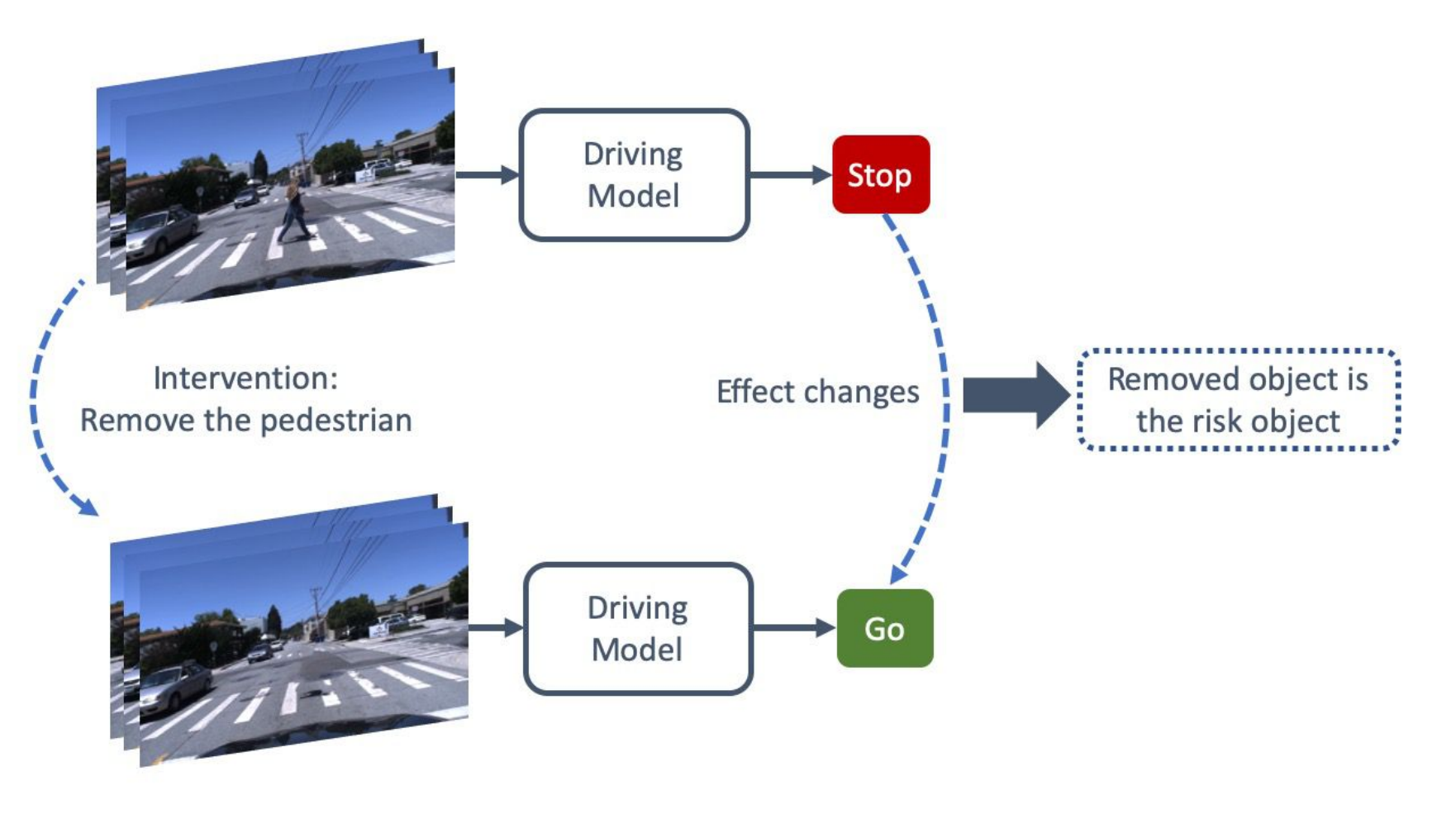}
                \caption{\textbf{Counterfactual intervention to measure the causal impact of an input region}. Removing a pedestrian \updated{through an object-level manipulable driving model} induces a change in the driver's decision from \emph{Stop} to \emph{Go}, which indicates that the pedestrian is a risk-object\updated{, \ie the cause for \emph{Stop}}.
                \updated{} 
                Credits to \citep{whomakedriversstop}.}
                \label{fig:sec:4:whomakesdriverstop}
            \end{figure}
        
            \indriving{}
            Regarding the autonomous driving literature, there only exists a limited number of approaches involving counterfactual interventions.
            When the input space has semantic dimensions and can thus be easily manipulated, it is easy to check for the causality of input factors by intervening on them (removing or adding).
            For example, \citet{chauffeurnet} investigate the causal factors for specific outputs: they test the ChauffeurNet model under hand-designed inputs where some objects have been removed.
            With a high-dimensional input space (\eg pixels), \citet{explaining_pilotnet} propose to check the causal effect that image parts have, with a saliency visualization method. In particular, they measure the effect of \emph{shifting} the image regions that were found salient by VisualBackProp on the PilotNet architecture. They observe that translating only these image regions, while maintaining the position of other non-salient pixels, leads to a significant change in the steering angle output. Moreover, translating non-salient image regions, while maintaining salient ones, leads to almost no change for the output of PilotNet. This analysis indicates a causal effect of the salient image regions.
            
            More recently, \citet{whomakedriversstop} introduce a causal inference strategy for the identification of ``risk-objects'', \ie objects that have a causal impact on the driver's behavior (see \autoref{fig:sec:4:whomakesdriverstop}).
            The task is formalized with an FCM and objects are removed in the input stream to simulate causal effects, the underlying idea being that removing non-causal objects will not affect the behavior of ego vehicles. Under this setting, they do not require strong supervision about the localization of risk-objects, but only the high-level behavior label (`go' or `stop'), as provided in the HDD dataset \citep{RamanishkaCMS18} for example.
            They propose a training algorithm with interventions, where some objects are randomly removed in scenes where the output is `go'. The object removal is instantiated with partial convolutions \citep{partial_convolutions}. At inference, in a sequence where the car predicts `stop', the risk-object is found as the one which gives the higher score to the `go' class.
            Finally, the model STEEX \citep{steex} presents a counterfactual method to explain deep visual models.
            In their case, the driving model is highly simplified and instantiated with a binary classifier \emph{Stop}/\emph{Move Forward} that operates on images taken by a frontal camera. 
            With the help of a pretrained generative model, STEEX generates counterfactual explanations where the scene structure is preserved and only the style of each region is allowed to be modified. Moreover, their architecture can generate counterfactual explanation that target a user-specified region of the query image. Results are shown in \autoref{fig:sec:4:steex}.

            \challenges{}
            We call the reader's attention to the fact that analyzing driving scenes and building driving models using causality is far from trivial as it requires the capacity to \emph{intervene} on the model's inputs. This, in the context of driving, is a highly complex problem to solve for three main reasons. 
            
            First, the data is composed of high-dimensional tensors of raw sensor inputs (such as the camera or LiDAR signals) and scalar-valued signals that represent the current physical state of the vehicle (velocity, yaw rate, acceleration, \textit{etc}.). Performing controlled interventions on these input spaces require the capacity to modify the content of raw high-dimensional inputs (\eg videos) realistically: changes in the input space such that counterfactual examples still belong to the data distribution, without producing meaningless perturbations alike adversarial ones.
            Even though some recent works explore realistic alterations of visual content \citep{flow_edge_guided}, this is yet to be applied in the context of self-driving and this open challenge, shared by other interpretability methods, is discussed in more details in \autoref{sec:5:input:semantic}.
            Interestingly, as more and more neural driving systems rely on semantic representations, alterations of the input space are simplified as the realism requirement is removed, and synthetic examples can be passed to the model as it has been done in \citep{chauffeurnet}.
            
            Second, modified inputs must be coherent and respect the underlying causal structure of the data generation process. Indeed, the different variables that constitute the input space are interdependant, and performing an intervention on one of these variables implies that we can simulate accordingly the reaction of other variables. As an example, we may be provided with a driving scene that depicts a green light, pedestrians waiting and vehicles passing. A simple intervention consisting of changing the state of the light to red would imply massive changes on the other variables to be \emph{coherent}: pedestrians should start crossing the street and vehicles should stop at the red light. The very recent and promising work of \citet{li2020causal} tackles the issue of unsupervised \emph{causal discovery} in videos. They discover a structural causal model in the form of a graph that describes the relational dependencies between variables. Interestingly, this causal graph can be leveraged to perform interventions on the data (\eg specify the state of one of the variables), leading to an evolution of the system that is coherent with this inferred graph. We believe that the adaptation of this type of approach to real driving data is crucial for the development of causal explainability.
            
            Finally, even if we are able to perform realistic and coherent interventions on the input space, we would need to have annotations for these new examples. Indeed, whether we use those altered examples to train a driving model on or to perform exhaustive and controlled evaluations, expert annotations would be required. Considering the nature of the driving data, it might be hard for a human to provide these annotations: they would need to imagine the decision they would have taken (control values or future trajectory) in this newly generated situation.

    \subsection{Global explanations}
        \label{sec:4:global}

        Global explanations contrast with local explanation methods as they attempt to explain the behavior of a model in general by summarizing the information it contains.
        \updated{They aim at increasing global and holistic \emph{model interpretability.}
        }
        We cover \updated{two} families of methods to provide global explanations: \emph{model translation} techniques, which aim at transforming an opaque neural network into a more interpretable model (\autoref{sec:4:global:translation}) and \emph{representations explanation} which analyze the knowledge contained in the data structures of the model (\autoref{sec:4:global:representations}). 

        \subsubsection{Model translation}
            \label{sec:4:global:translation}
 
            \ingeneral{}
            The idea of \emph{model translation} is to transfer the knowledge contained in the main opaque model into a separate machine learning model that is inherently interpretable.
            Concretely, this involves training an explainable model to mimic the input-output mapping of the black-box function. Despite sharing the same spirit with local approximation methods presented in \autoref{sec:4:local:saliency}, model translation methods are different as they should approximate the main function \emph{globally} across the data distribution.
            In the work of \citet{cnn_knowledge_explanatory_graph} an explanatory graph is built from a pre-trained convolutional neural net to understand how the patterns memorized by its filters are related to object parts. This graph aims at providing a global view of how visual knowledge is organized within the hierarchy of convolutional layers in the network. Deep neural networks have also been translated into soft decision trees \citep{soft_decision_tree} or rule-based systems \citep{deepred,sato2001rule}.
            The recent work of \citet{causal_learning_autoencoded_activations} presents a causal model used to explain the computation of a deep neural network. Human-understandable concepts are first extracted from the neural network of interest, using auto-encoders with sparsity losses. Then, the causal model is built using those discovered human-understandable concepts and can quantify the effect of each concept on the network's output.

            \challenges{}
            To the best of our knowledge, such strategies have not been used in the autonomous driving literature to visualize and interpret the rules learned by a neural driving system. 
            Indeed, one of the limit of such a strategy lies in the disagreements between the interpretable translated model and the main self-driving model. These disagreements are inevitable as rule-based models or soft-decision trees have a lower capacity than deep neural networks.
            Moreover, these methods are typically designed to explain deep networks that perform a classification task, which is usually not the case of self-driving models. \updated{Being able to adapt these techniques to driving models could certainly be beneficial to engineers and regulatory institutions. Indeed, these methods would provide a direct understanding on the knowledge captured by learning-based algorithms, which could help assess their safety and conformity to societal rules and ethics.}

        \subsubsection{Explaining representations}
            \label{sec:4:global:representations}

            \ingeneral{}
            Representations in deep networks take various forms as they are organized in a hierarchy that encompasses individual units (neuron activation), vectors, and layers \citep{GilpinBYBSK18}.
            The aim of explaining representations is to provide insights into what is captured by the internal data structures of the model, at different granularities.
            Representations are of practical importance in transfer learning scenarios, \ie when they are extracted from a deep network trained on a task and transferred to bootstrap the training of a new network optimizing a different task. %
            
            In practice, the quality of intermediate representations can be evaluated, and thus made partially interpretable, with a proxy transfer learning task \citep{cnn_features_off_the_shelf}.
            At another scale, some works attempt to gain insights into what is captured at the level of an individual neuron \citep{network_dissection,visual_interpretability_dl_survey}.
            For example, a neuron's  activation can be interpreted by accessing input patterns which maximize its activation, for example by sampling such input images \citep{object_detectors_emerge_in_deep_scene_cnns,learning_aligned_cross_modal_representations}, with gradient ascent \citep{visualizing_higher_layer_features,deep_inside_cnn}, or with a generative network \citep{synthesizing_preferred_inputs_neurons}.
            To gain more understanding of the content of vector activations, the t-Distributed Stochastic Neighbor Embedding (t-SNE) \citep{tsne} has been proposed to project high-dimensional data into a space of lower dimension (usually 2d or 3d).
            This algorithm aims at preserving the distances between points in the new space where points are projected. 
            t-SNE has been widely employed to visualize and gain more interpretability from representations, by producing \emph{scatter plots} as explanations.
            This has for example been employed for video representations \citep{C3D}, or deep Q-networks \citep{graying_black_box}.
            
            \challenges{}
            In the autonomous driving literature, such approaches have not been widely used to the best of our knowledge. The only example we can find is reported in \citep{deeptest} which uses the neuron \emph{coverage} concept from \citep{deepxplore}. The neuron coverage is a testing metric for deep networks, that estimates the amount of logic explored by a set of test inputs: more formally the neuron coverage of a set of test inputs is the proportion of unique activated neurons, among all network's neurons for all test inputs. %
            \updated{With the idea of detecting erroneous behaviors of deep self-driving models that could lead to potential accidents, \citet{deeptest} partition the input space according to the neuron coverage by assuming that the model decision is the same for inputs that have the same neuron coverage. With the aim of increasing neuron coverage of the model, they compose a variety of transformation of the input image stream, each corresponding to a synthetic but realistic editing of the scene: linear (\eg change of luminosity/contrast), affine (\eg camera rotation) and convolutional (\eg rain or fog) transformations. This enables them to automatically discover many --- synthetic but realistic --- scenarios where the predictions are incorrect. Interestingly, they show that the insights obtained on erroneous corner cases can be leveraged to successfully retrain the driving model on the synthetic data to obtain an accuracy boost. Despite not giving explicit explanations about the self-driving model, such predictions help to understand the model's limitations. 
            }
            \updated{Overall, inspecting neurons and intermediate representations is a tedious and lengthy operation. The detailed information provided by this inspection can be very precious to machine learning experts, but certainly not to regular car users.}

\begin{table*}[t]
    \centering
    \rowcolors{2}{white}{white}
    \begin{tabular}{@{}l l l p{7.5cm} @{}}
        \toprule 
        Approach & Explanation type & Section & Selected references \\ %
        \hline 
         & Attention maps & \ref{sec:5:input:attention} &
        \makecell[l]{Visual attention \citep{causal_attention} \\ Object centric \citep{deep_object_centric} \\ Attentional Bottleneck \citep{attentional_bottleneck}} \\
        \multirow{-5}{*}{Input interpretability} & Semantic inputs & \ref{sec:5:input:semantic} & \makecell[l]{DESIRE \citep{desire} \\ ChauffeurNet \citep{chauffeurnet} \\ MTP \citep{uncertainty_aware_short_term_motion_prediction,mtp}} \\
        \midrule
         & Auxiliary branch & \ref{sec:5:intermediate:1} & \makecell[l]{
        Affordances/action primitives \citep{guidedsupervision} \\ 
        Detection/forecast of vehicles \citep{neural_motion_planner} \\ 
        Multiple auxiliary losses \citep{chauffeurnet}} \\ 
        \multirow{-4}{*}{Intermediate representations} & Output & \ref{sec:5:intermediate:1} & \makecell[l]{
        Auto-regressive likelihood map \\ \citep{SrikanthARSMK19,chauffeurnet} \\
        Segmentation of future track in \bev{} \\ \citep{lidar_based_driving} \\
        Cost-volume \citep{neural_motion_planner} \\
        Sequences of points \citep{desire} \\
        Classes \citep{covernet}} \\
        
        & NLP & \ref{sec:5:intermediate:2} & \makecell[l]{Natural language \citep{textual_explanations,attention_branch_network}} \\
        \bottomrule 
    \end{tabular}
    \caption{\textbf{Selected references to design an explainable driving model}.
    }
    \label{tab:explainable_design}
\end{table*}

\section{Designing an explainable driving model}
    \label{sec:5}
    
    In the previous section, we saw that it is possible to explain the behavior of a machine learning model locally or globally, using post-hoc tools that make little to no assumption about the model. 
    Interestingly, these tools operate on models whose design may have completely ignored the requirement of explainability. 
    A good example of such models is PilotNet \citep{pilotnet,pilotnet2}, which consists in a convolutional neural network operating over a raw video stream and producing the vehicle controls at every time step. 
    Understanding the behavior of this system is only possible through external tools, such as the ones presented in \autoref{sec:4}, but cannot be done directly by observing the model itself.
    
    \updated{On the other hand, a recent trend propose to integrate explainability constraints directly \emph{during} the design of deep architectures such that they exhibit increased interpretability levels \citep{interpretable_cnn,this_looks_like_that,interpretable_categorization_time_series}}.
    Drawing inspiration from modular systems, recent architectures place a particular emphasis on conveying understandable information about their inner workings, in addition to their performance imperatives.
    As was advocated in \citep{explainable_object_induced}, the modularity of pipelined architectures allows for forensic analysis, by studying the quantities that are transferred between modules (\eg semantic and depth maps, forecasts of surrounding agent's future trajectories, \textit{etc}.). 
    Moreover, finding the right balance between modular and end-to-end systems can encourage the use of simulation, for example by training separately perception and driving modules \citep{driving_policy_transfer}.
    
    These modularity-inspired models exhibit some forms of interpretability, which can be enforced at three different levels in the design of the driving system.
    \updatedlast{ 
    We first review \emph{input-level explanations} (\autoref{sec:5:input}), which aim at communicating which perceptual information is used by the model. 
    We then study \emph{intermediate-level explanations} (\autoref{sec:5:intermediate}) which force the network to produce supplementary information as it drives.
    }
    Selected references from this section are reported in \autoref{tab:explainable_design}.

    \subsection{Input} 
        \label{sec:5:input}
        Input-level explanations aim at enlightening the user on which perceptual information is used by the model to take its decisions.
        We identified two families of approaches that ease interpretation at the input level: attention-based models (\autoref{sec:5:input:attention}) and models that use semantic inputs (\autoref{sec:5:input:semantic}).

        \subsubsection{Attention-based models}
            \label{sec:5:input:attention}

            \ingeneral{}
            Attention mechanisms, initially designed for NLP application \citep{attention_bahdanau}, learn a function that scores different regions of the input depending on whether or not they should be considered in the decision process. 
            This scoring is often performed based on some contextual information that helps the model decide which part of the input is relevant to the task at hand. 
            \citet{show_attend_tell} are the first to use an attention mechanism for a computer vision problem, namely, image captioning. In this work, the attention mechanism uses the internal state of the language decoder to condition the visual masking. The network knows which words have already been decoded, and seeks for the next relevant information inside of the image.
            Many of such attention models were developed for other applications since then, for example in Visual Question Answering (VQA) \citep{ask_attend_answer,coattention_vqa,stacked_attention_question_answering}. These systems, designed to answer questions about images, use a representation of the question as a context to the visual attention module. Intuitively, the question tells the VQA model where to look to answer the question correctly.
            Not only do attention mechanisms boost the performance of machine learning models, but also they provide insights into the inner workings of the system. Indeed, by visualizing the attention weight associated with each input region, it is possible to know which part of the image was deemed relevant to make the decision. %

            \indriving{}
            Attention-based models recently stimulated interest in the self-driving community, as they supposedly give a hint about the internal reasoning of the neural network. 
            In \citep{causal_attention}, an attention mechanism is used to weight each region of an image, using information about previous frames as a context. 
            A different version of attention mechanisms is used in \citep{attention_branch_network}, where the model outputs a steering angle and a throttle command prediction for each region of the image. These local predictions are used as attention maps for visualization and are combined through a linear combination with learned parameters to provide the final decision.
            
\begin{figure}
                \centering
                \includegraphics[width=\linewidth]{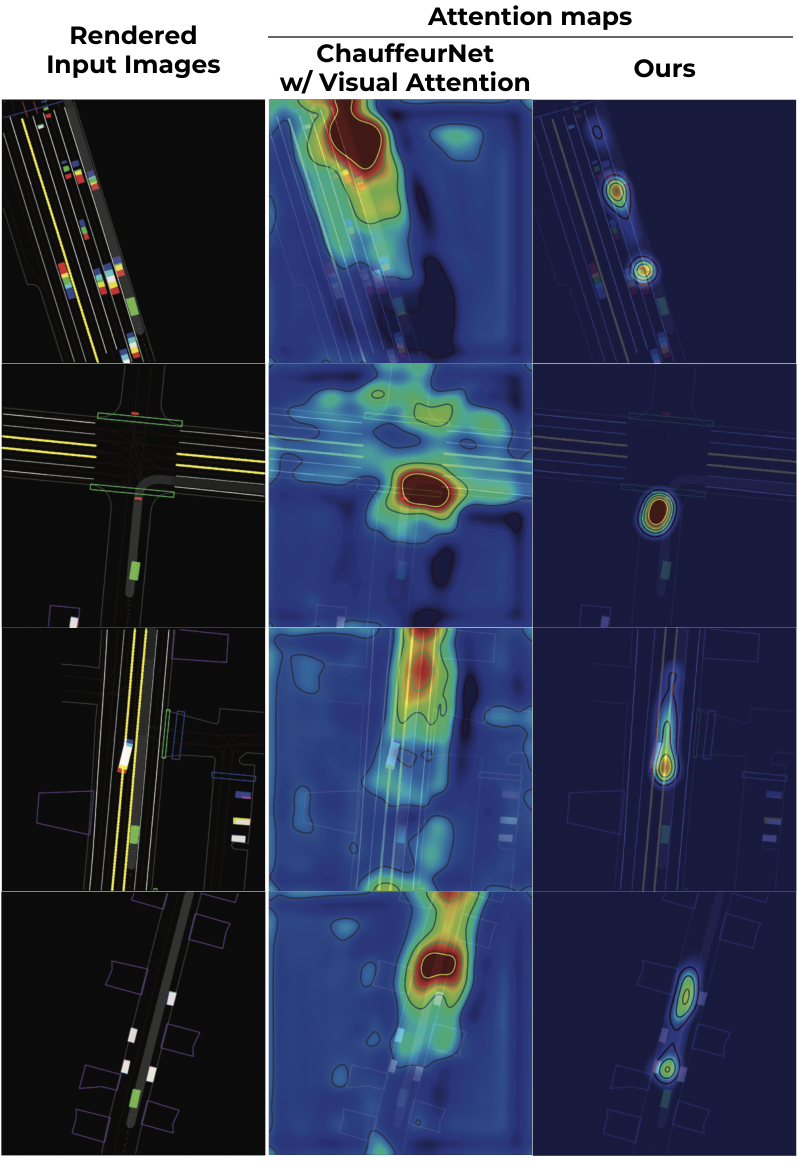}
                \caption{\updated{
                \textbf{Comparison of attention maps} with classical visual attention (middle column) and maps produced by the attentional bottleneck method \citep{attentional_bottleneck} (right column).
                Attentional bottleneck provides sparser saliency maps with tighter modes that focus on objects of interest such as surrounding cars. 
                }
                Credits to \citep{attentional_bottleneck}.}
                \label{fig:sec:5:attentional_bottleneck}
            \end{figure}

            Visual attention can also be used to select objects defined by bounding boxes, as in \citep{deep_object_centric}. In this work, a pre-trained object detector provides regions of interest (RoIs), which are weighted using the global visual context, and aggregated to decide which action to take; their approach is validated on both simulated GTAV \citep{gtav} and real-world BDDV \citep{bddv} datasets.
            \citet{cultrera2020} also use attention on RoIs in a slightly different setup with the CARLA simulator \citep{carla}, as they directly predict a steering angle instead of a high-level action.
            Recently, \citet{attentional_bottleneck} extended the ChauffeurNet \citep{chauffeurnet} architecture by building a visual attention module that operates on a bird-eye view semantic scene representation. Interestingly, as shown in \autoref{fig:sec:5:attentional_bottleneck}, combining visual attention with information bottleneck results in sparser saliency maps, making them more interpretable. %
            
            \challenges{}
            While these attention mechanisms are often thought to make neural networks more transparent, the recent work of \citet{attention_not_explanation} mitigates this assumption. Indeed, they show, in the context of natural language, that learned attention weights poorly correlate with multiple measures of feature importance. Moreover, they show that randomly permuting the attention weights usually does not change the outcome of the model. They even show that it is possible to find adversarial attention weights that keep the same prediction while weighting the input words very differently. Even though some works attempt to tackle these issues by learning to align attention weights with gradient-based explanations \citep{PatroAN20}, all these findings cast some doubts on the faithfulness of explanations based on attention maps. \updated{These limitations restrict the use of attention maps for engineers and legal institutions, who have high requirements for fidelity (see \autoref{sec:2:xai_context}). However, similarly to post-hoc visual saliency methods presented in \autoref{sec:4:local:saliency}, \updated{interpretable-by-design} attention maps provide acceptable explanations to increase the trust of regular car users, as they provide an instantaneous glimpse into the inner workings of the model.}

        \subsubsection{Semantic inputs}
            \label{sec:5:input:semantic}
            
            Some traditional machine learning models such as linear and logistic regressions, decision trees, or generalized additive models are considered interpretable by practitioners \citep{molnar2019}. 
            However, as was remarked by \citet{self_explaining_nn}, these models tend to consider each input dimension as the fundamental unit on which explanations are built. 
            Consequently, the input space must have a semantic nature such that explanations become interpretable.
            Intuitively, each input dimension should \emph{mean something} independently of other dimensions.
            In general machine learning, this condition is often met, for example with categorical and tabular data.
            However, in computer vision, when, dealing with images, videos, and 3D point clouds, the input space has not an interpretable structure.
            Overall, in self-driving systems, the lack of semantic nature of inputs impacts the interpretability of machine learning systems.
            
            This observation has motivated researchers to design, build, and use more interpretable input spaces, for example by enforcing more structure or by imposing dimensions to have an underlying high-level meaning. 
            The promise of a more interpretable input space towards increased explainability is diverse.
            First, the visualization of the network's attention or \emph{post-hoc} saliency heat maps in a semantic input space is more interpretable as it does not apply to individual pixels but rather to higher-level object representations. \updated{In practice, visualizing input information in a semantic space can improve the trust of regular car users in the driving model.}
            Second, counterfactual analysis, \updated{useful to regulators and engineers (cf. \autoref{sec:2:xai_context}),} is simplified as the input can be manipulated more easily without the risk of generating meaningless imperceptible perturbations, akin to adversarial attacks.

            \paragraph{Using semantic inputs.}
            \label{sec:5:input:semantic:using}

            Apart from camera inputs processed with deep CNNs in \citep{pilotnet,codevilla2018end}, different approaches have been developed to use semantic inputs in a self-driving model, depending on the types of signals at hand.
            
            3D point clouds, provided by LiDAR sensors, can be processed to form a top-view representation of the car surroundings. For instance, \citet{lidar_based_driving} propose to flatten the scene along the vertical dimension to form a top-down map, where each pixel in the \bev{} corresponds to a 10cm$\times$10cm square of the environment.
            While this input representation provides information about the presence or absence of an obstacle at a certain location, it crucially lacks semantics as it ignores the nature of the obstacles (sidewalks, cars, pedestrians, \textit{etc.}).
            This lack of high-level scene information is attenuated in DESIRE \citep{desire}, where the output of an image semantic segmentation model is projected to obtain labels in the top-down view generated from the LiDAR point cloud. In DESIRE, static scene components are projected within the top-down view image (\eg road, sidewalk, vegetation), and moving agents are represented along with their tracked present and past positions.
            
\begin{figure}
                \centering
                \includegraphics[trim={35 20 5 5},clip,width=0.8\linewidth]{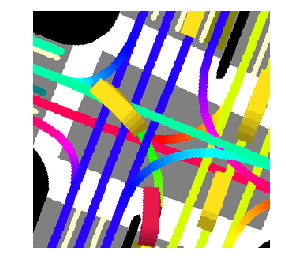}
                \caption{
                \updated{ 
                \textbf{RGB \bev{}}. Semantic information is rasterized within an RGB image that constitutes the input of a convolutional driving backbone.
                For example, surrounding cars are represented as yellow rectangles and their previous positions are also depicted with a shade.
                The lanes are encoded with colors corresponding to their orientation.
                } 
                Credits to \citep{uncertainty_aware_short_term_motion_prediction}.}
                \label{fig:sec:5:uncertainty_aware}
            \end{figure}

            The ChauffeurNet model \citep{chauffeurnet} relies on a similar top-down scene representation, however instead of originating from a LiDAR point cloud, the \bev{} is obtained from city map data (such as speed limits, lane positions, and crosswalks), traffic light state recognition and detection of surrounding cars. These diverse inputs of the network are gathered into a stack of several images, where each channel corresponds to a rendering of a specific semantic attribute. 
            This contrasts with more recent approaches that aggregate all information into a single RGB top-view image, where different semantic components correspond to different color channels \citep{uncertainty_aware_short_term_motion_prediction,mtp}. While the information is still semantic, having a 3-channel RGB image allows leveraging the power of pre-trained convolutional networks. An example RGB semantic image is shown in \autoref{fig:sec:5:uncertainty_aware}. %
            
            \paragraph{Towards more control on the input space.}
            
         \begin{figure*}
            \centering
            \includegraphics[trim={0 25 10 0},clip,width=0.9\linewidth]{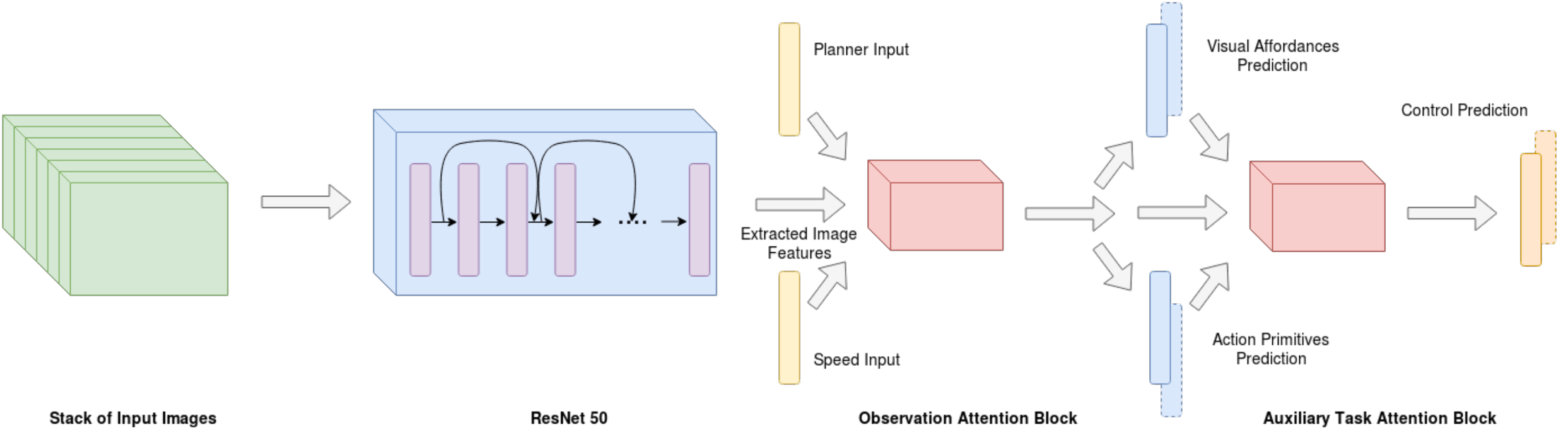}
            \caption{\textbf{Predicting intermediate representations}.
            \updated{
            The end-to-end driving model operates over stacked images as well as inputs from the planner to produce control commands.
            Auxiliary objectives are employed for the joint prediction of visual affordances (abstract description of the visual scene) and action primitives (abstract description of possible high-level actions required for driving).
            }
            Credits to \citep{guidedsupervision}.
            }
            \label{fig:sec:5:mehta}
        \end{figure*}

            Having a manipulable input space where we can play on semantic dimensions (\eg controlling objects' attributes, changing the weather, removing a specific car) is a very desirable feature for increased explainability of self-driving models.
            Importantly, having such a feature would nicely synergies with many of the post-hoc explainability methods presented in \autoref{sec:4}. For example, to produce counterfactual examples, without falling into adversarial meaningless perturbations, it is desirable to have an input space on which we can apply semantic modifications at a pixel-level. As other examples, local approximation methods such as LIME \citep{lime} would highly benefit from having a controllable input space as a way to ease the sampling of locally similar scenes.

            Manipulating inputs can be done at different semantic levels.
            First, at a global level, changes can include the scene lighting (night/day) and the weather (sun/rain/fog/snow) of the driving scene \citep{deeptest}, and more generally any change that separately treats style and texture from content and semantics \citep{towards_photo_realistic_facial_expression} ; such global changes can been done with video translation models \citep{mocogan,recycle_gan,optical_flow_distillation}.
            At a more local level, possible modifications include adding or removing objects \citep{whomakedriversstop,vornet,recovering_and_simulating_pedestrians}, or changing attributes of some objects \citep{fadernet}.
            Recent video inpainting works \citep{flow_edge_guided} can be used to remove objects from videos.
            Finally, at an intermediate level, we can think of other semantic changes to be applied to images, such as controlling the proportion of classes in an image \citep{layout2image,semantic_palette}.
            Manipulations could be done by playing on attributes \citep{fadernet}, by inserting virtual objects in real scenes \citep{augmented_reality_meets_computer_vision,scenegen}, \updated{by augmenting existing images with dynamic objects extracted from other scenes \citep{geosim}}, or by the use of textual inputs with GANs \citep{manigan,lightweight_gan_text_guided}.

            We note that having a semantically controllable input space can have lots of implications for areas connected with interpretability. For example, to validate models, and towards having a framework to certify models, we can have a fine-grain stratified evaluation of self-driving models. This can also be used to automatically find failures and corner cases by easing the task of exploring the input space with manipulable inputs \citep{deeptest}. Finally, to aim for more robust models, we can even use these augmented input spaces to train more robust models, as a way of data augmentation with synthetically generated data \citep{gan_augmentation,red_blood_cell}.

    \subsection{Supervising intermediate representations} 
        \label{sec:5:intermediate}
    
    \updatedlast{A neural network makes its decisions by automatically constructing intermediate representations of the data. 
    These intermediate representations can be constrained to encode for extra driving information such as underlying high-level objectives of the driving system (\autoref{sec:5:intermediate:1}), to mimic explanations that humans would provide in similar situations (\autoref{sec:5:intermediate:2}). Without explicit constraint, these representations can still be visualized or disentangled for more interpretability (\autoref{sec:5:intermediate:3}).
    }

    \subsubsection{\updatedlast{Providing auxiliary driving information}}
    \label{sec:5:intermediate:1}
            
            \updatedlast{
            The task of autonomous driving consists in continuously producing the suitable vehicle commands, \ie steering angle, brake, and throttle controls. 
            A very appealing solution is to train a neural network to directly predict these values \citep{DBLP:conf/nips/Pomerleau88,pilotnet,codevilla2018end}. 
            However, having a system that directly predicts these command values may not be satisfactory in terms of interpretability, as it may fail to communicate the precise understanding of the surroundings from the perception models, as well as local objectives that the vehicle is attempting to attain.
            }

            One way of creating interpretable driving models is to enforce that some information, different than the one directly needed for driving, is present in these features. 
            Doing so, the prediction of a driving decision can be accompanied by an auxiliary output that provides a human-understandable view of the information contained in the intermediate features. \updated{Moreover, as was stated in \citep{doescomputervisionmatterforaction}, sensorimotor agents benefit, in terms of accuracy, from predicting explicit intermediate scene representations in parallel to their main task.}
            In \citep{guidedsupervision}, as illustrated in \autoref{fig:sec:5:mehta}, a neural network learns to predict control outputs from input images. Its training is helped with auxiliary tasks that aim at recognizing high-level action primitives (\eg ``stop'', ``slow down'', ``turn left'', \textit{etc}.) and visual affordances in the CARLA simulator \citep{carla}. %
             \begin{figure}
                \centering
                \includegraphics[width=\columnwidth]{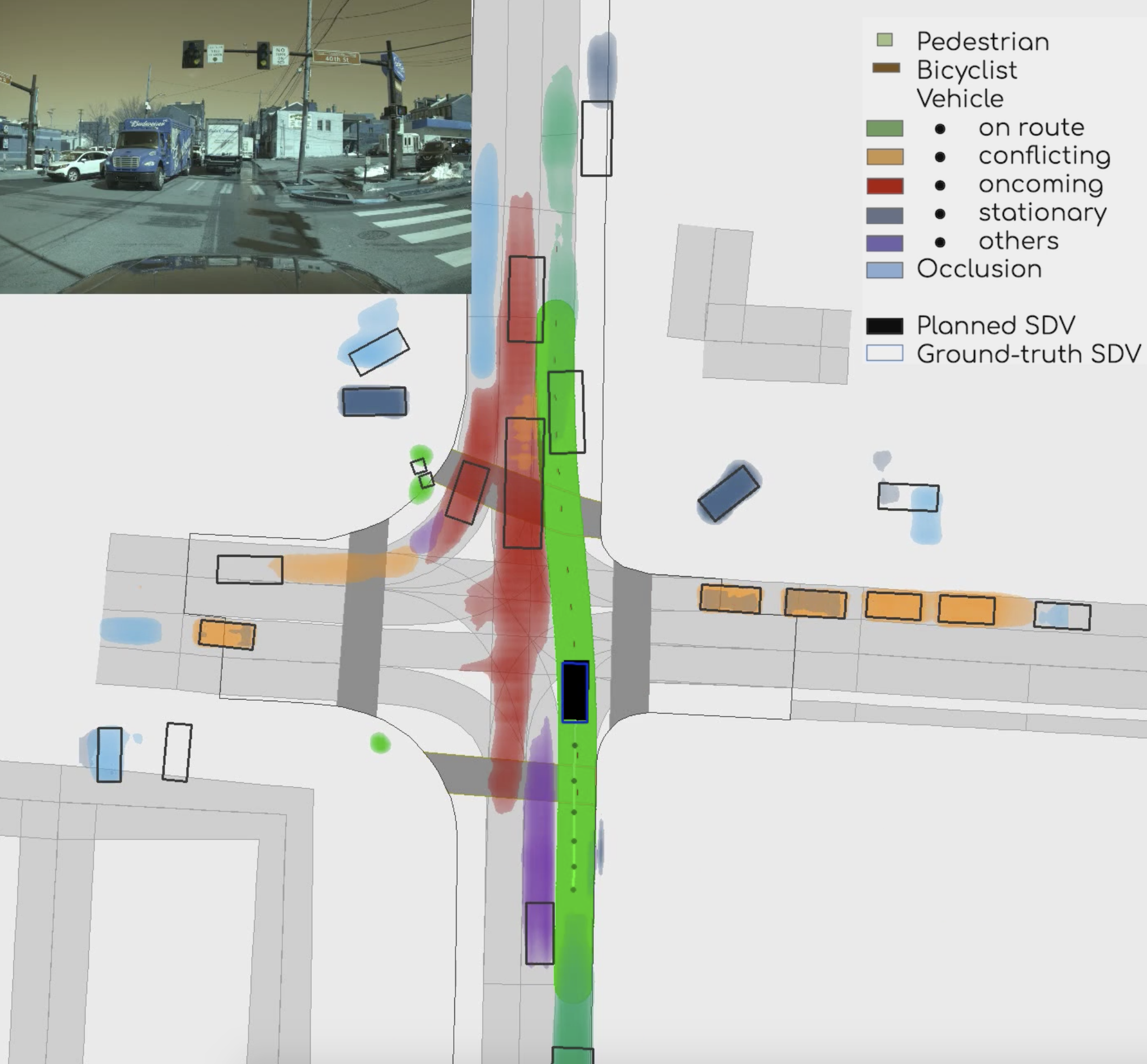}
                \caption{
                \updated{
                \textbf{Semantic occupancy maps}.
                This semantic map is predicted with an auxiliary objective while training the end-to-end motion planner. Colors represent the probability of the future occupancy of various semantic classes (pedestrian, bicycles, vehicles, etc.)
                Credits to \citep{perceive_predict_plan}.
                }
                }
                \label{fig:sec:5:perceive_predict_plan}
            \end{figure}

            \begin{figure*}
                \centering
                \includegraphics[width=0.8\linewidth]{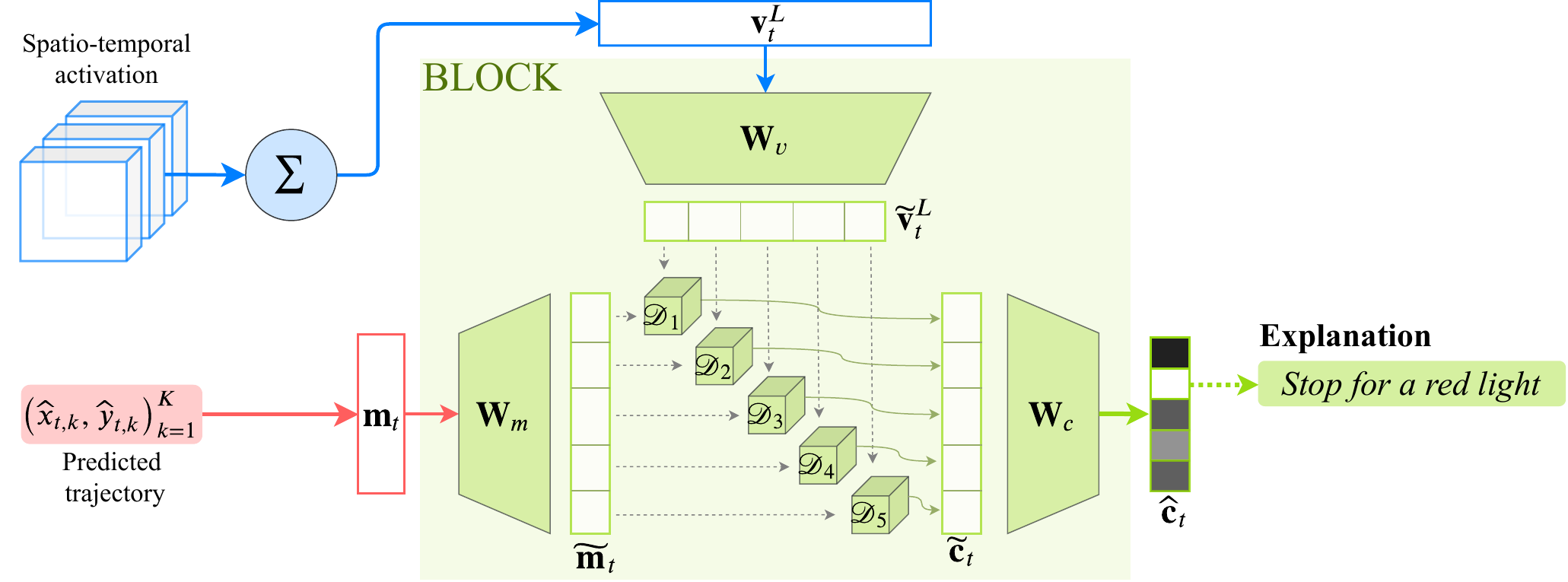}
                \caption{\textbf{Multi-level fusion for explanation prediction}. The explanation \updated{(in green)} for a driving decision is expressed as a fusion between the predicted trajectory\updated{, \ie the driving output (in red)} and perceptual features\updated{, \ie mid-level features from the backbone (in blue)
                 that bring complementary information}. Credits to \citep{beef}.}
                \label{fig:sec:5:beef}
            \end{figure*}

            In \citep{neural_motion_planner}, a neural network predicts the future trajectory of the ego-vehicle using a top-view \lidar{} point-cloud. In parallel to this main objective, they learn to produce an interpretable intermediate representation composed of 3D detections and future trajectory predictions. Multi-task in self-driving has been explored deeply in \citep{chauffeurnet}, where the authors design a system with ten losses that, besides learning to drive, also forces internal representations to contain information about on-road/off-road zones and future positions of other objects.
            \updated{
            In \citep{perceive_predict_plan}, an end-to-end motion planning model produces interpretable intermediate representations in the form of probabilistic semantic occupancy maps over space and time.
            These spatio-temporal maps express the future locations of objects of different classes, as illustrated in \autoref{fig:sec:5:perceive_predict_plan}, in a probabilistic and instance-free fashion. 
            Their goal is to capture whether and when a discretized spatial region is occupied by an agent at a certain time step.
            These maps have the potential to remove the need for both detection and tracking while offering similar levels of interpretability.
            }
            
        \updatedlast{
        Most of recent models do not directly output driving commands, but rather some form of intermediate goals that the model should reach. Those are then passed to a controller that finds the suitable steering, brake and acceleration commands to reach the required position. 
        The intermediate goals can be visualized in the same coordinate system as the input representation, which helps the human user interpret the prediction with respect to scene elements (road structure, surrounding agents, \textit{etc}.).
        }
        
        \updatedlast{ 
        Output representations of neural trajectory prediction systems can be split into two categories: analytical representations and spatial grid representations. 
        }
        
        \updatedlast{
        Systems that output an analytical representation of the future trajectory provide one or more predictions in the form of points or curves in the 2D space. 
        This output structure is commonly used in motion forecasting models \citep{desire,mtp,covernet,trajectron++,cab}. 
        For instance, \citet{desire} propose to predict multiple future trajectories for each agent of the scene. More specifically, recurrent models are trained to sample trajectories as sequences of 2D points in a bird's-eye-view basis, rank them, and refine them according to perceptual features. %
        In MTP \citep{mtp},
        each trajectory consists of a set of 2D points and a confidence score. %
        CoverNet \citep{covernet} poses the trajectory prediction problem as a classification one, where each possible class is a predefined trajectory profile. %
        }

        \updatedlast{
        In the second family of trajectory prediction systems, the network scores regions of the spatial grid according to their likelihood of hosting the car in the future. %
        In \citep{lidar_based_driving}, the network is trained to predict the track of the future positions of the vehicle, in a semantic segmentation fashion. %
        Differently, ChauffeurNet \citep{chauffeurnet} predicts the next vehicle position as a probability distribution over the spatial coordinates. 
        The Neural Motion Planner \citep{neural_motion_planner} contains a neural network that outputs a cost volume, which is a spatio-temporal quantity indicating the cost for the vehicle to reach a certain position at a certain moment. Trajectories are sampled from a set of dynamically possible paths (straight lines, circles, and clothoïds) and scored according to the cost volume. Interestingly, the cost volume can be visualized, and thus provides a human-understandable view of what the system considers feasible.
        }

    \subsubsection{\updatedlast{Using human explanations}}
    \label{sec:5:intermediate:2}

            Instead of supervising intermediate representations with scene information, other approaches propose to directly use explanation annotations as an auxiliary branch. The driving model is trained to simultaneously decide \emph{and} explain its behavior.
            In the work of \citet{explainable_object_induced}, the BDD-OIA dataset was introduced, where clips are manually annotated with \emph{authorized} actions and their associated explanation. Action and explanation predictions are expressed as multi-label classification problems, which means that multiple actions and explanations are possible for a single example. 
            While this system is not properly a driving model (no control or trajectory prediction here, but only high-level classes such as ``stop'', ``move forward'' or ``turn left''), \citet{explainable_object_induced} were able to increase the performance of action decision making by learning to predict explanations as well. 
            
            Very recently, \citet{beef} propose to explain the behavior of a driving system by fusing high-level decisions with mid-level perceptual features. The fusion, depicted in \autoref{fig:sec:5:beef}, is performed using BLOCK \citep{ben2019block}, a tensor-based fusion technique designed to model rich interactions between heterogeneous features. 
            Their model is trained on the HDD dataset \citep{RamanishkaCMS18}, where 104 hours of human driving are annotated with a focus on driver behavior. In this dataset, video segments are manually labeled with classes that describe the goal of the driver (\eg ``turn left'', ``turn right'', \textit{etc.}) as well as an explanation for its stops and deviations (\eg ``stop for a red light'', ``deviate for a parked car'', \textit{etc}).
            The architecture of \citet{beef} is initially developed to provide explanations in a classification setup, and they show an extension of it to generate natural language sentences (see \autoref{sec:6:nle}).

    \subsubsection{\updatedlast{Visualizing and disentangling representations}}
    \label{sec:5:intermediate:3}
    
            Visualizing the predictions of an auxiliary head is an interesting way to give the human user an idea of what information is contained in the intermediate representation. 
            Indeed, observing that internal representations of the driving network can be used to recognize drivable areas, estimate pedestrian attributes \citep{detecting_pestrian_attributes}, detect other vehicles, and predict their future positions strengthens the trust one can give to a model. Yet, it is important to keep in mind that information contained in the representation is not necessarily used by the driving network to make its decision. More specifically, the fact that we can infer future positions of other vehicles from the intermediate representation does not mean that these forecasts were actually used to make the driving decision. Overall, one should be cautious about such auxiliary predictions to interpret the behavior of the driving model, as the causal link between these auxiliary predictions and the driving output is not enforced.
            
            \updated{Interestingly, models have been developed to learn and discover disentangled latent variables without using auxiliary supervision. These unsupervised methods aim at capturing underlying salient data factors, such that each individual variable represents a single interpretable attribute \citep{representation_learning_review,infoGAN}. While the application of these approaches remains scarce in autonomous driving \citep{simultaneous_policy_learning}, their popularity is rising in image synthesis as they offer a way to control the generation over human-understandable variation factors \citep{vae_images,stylegan}.}

\section{Use case: natural language explanations}
\label{sec:6}

    \begin{figure*}
        \centering
        \includegraphics[width=0.75\linewidth]{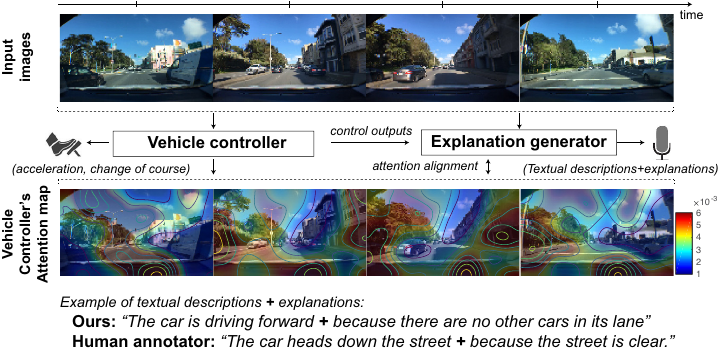}
        \caption{\textbf{Generating natural language explanations}. 
        \updated{
        The end-to-end driving model (`vehicle controller') predicts the control commands (acceleration and change of direction). The predicted control outputs are provided to the `Explanation generator' that also sees the visual scene. This module is trained to provide natural language justifications describing the scene and explaining the driving decision taken by the driving model.
        The explanation generator is forced to align its attention on the one learnt by the vehicle controller.
        }
        Credits to \citep{textual_explanations}.
        }
        \label{fig:sec:5:textual_explanations}
    \end{figure*}
            
As was stated in \autoref{sec:2:xai_context}, some of the main requirements of explanations targeted at non-technical human users are conciseness and clarity. 
To meet these needs, some research efforts have been geared at building models that provide explanations of their behavior in the form of natural language sentences. In \autoref{sec:6:nle}, we review the methods proposed by the community to generate natural language explanations of machine learning models. The limits of such techniques are discussed in \autoref{sec:6:limits}. %

\subsection{Generating natural language explanations.}
\label{sec:6:nle}

    \ingeneral{}
    The first attempt to explain the predictions of a deep network with natural language was in the context of image classification, where \citet{generating_visual_explanations} train a neural network to generate sentence explanations from image features and class label. These explanations are forced to be \textit{relevant} to the image, \ie to mention elements that are present in the image, and also \textit{class-discriminative}, which means they can spot specific visual elements that separate one class from another. This work is further extended in \citep{grounding_visual_explanations}, where a list of candidate explanations is sorted with respect to how noun phrases are visually-grounded.
    
    In the field of natural language processing (NLP), \citet{explainable_nlp} build an explanation-producing system for long review text classification. In particular, they tackle the problem of independence between the prediction and its explanation and try to strengthen the connection between both. To do so, they pre-train a classifier that takes as input an explanation and predicts the class of the associated text input, and they use this classifier to measure and optimize the difference between true and generated explanations.
    Moreover, \citet{esnli} propose to learn from human-provided explanations at train time for a natural language inference task. Similarly, \citet{explain_yourself} gather a dataset of human natural language explanations for a common-sense inference task and learn a model that jointly classifies the correct answer and generates the correct explanation.
    
    In the field of vision-and-language applications, \citet{multimodal_explanations} build ACT-X and VQA-X, two datasets of multi-modal explanations for the task of action recognition and visual question answering. More specifically, VQA-X (resp. ACT-X) contains textual explanations that justify the answer (resp. the action), as well as an image segmentation mask that shows areas that are relevant to answer the question (resp. recognize the action). Both textual and visual explanations are manually annotated. Related to this work, \citet{recognition_to_cognition} design a visual commonsense reasoning task where a question is asked about an image, and the answer is a sentence to choose among a set of candidates. Each example is also associated with another set of sentences containing candidate justifications of the answer and describing the reasoning behind a decision. 
        
    \indriving{}
    In the context of self-driving, \citet{textual_explanations} learn to produce textual explanations justifying decisions from a self-driving system. Based on the video material of BDDV \citep{bddv}, the authors built the BDD-X dataset where dash-cam video clips are annotated with a sentence that describes the driving decision (\eg ``\textit{the car is deviating from its main track}''), and another one that explains why this is happening (\eg ``\textit{because the yellow bus has stopped}'').
    An end-to-end driving system equipped with visual attention is first trained on this dataset to predict the vehicle controls for each frame, and, in a second phase, an attention-based video-to-text captioning model is trained to generate natural language explanations justifying the system's decisions. The attention of the captioning explanation module is constrained to align with the attention of the self-driving system. We show an overview of their system in \autoref{fig:sec:5:textual_explanations}. 
    Notably, this model is akin to a post-hoc explanation system as the explanation-producing network is trained after the driving model.
    
     The BDD-X dataset is also used by \citet{beef} as they adapt their explanation classification method to the setup of natural language generation. Interestingly, they study the impact of the temperature parameter in the decoding softmax, classically used to control the diversity of generated sentences, on the variability of sampled explanations for the same situation. In particular, they show that for reasonably low values of the temperature, the model justifies a driving situation with semantically consistent sentences. These explanations differ from each other only \emph{syntactically} and with respect to their \emph{completeness} (some explanations are more exhaustive and precise than others), but not \emph{semantically}. Looking at the example shown in \autoref{tab:sec:6:beef_temp}, we see that all the explanations are correct as they correspond to the depicted scene, but the level of detail they convey may be different.
     
    \begin{table}
        \centering
        \rowcolors{0}{white}{white}
        \resizebox{\linewidth}{!}{%
        \begin{tabular}{@{}l l@{}}
            \rotatebox{90}{\hspace{0.2cm} Extracted frame} & \includegraphics[height=30mm]{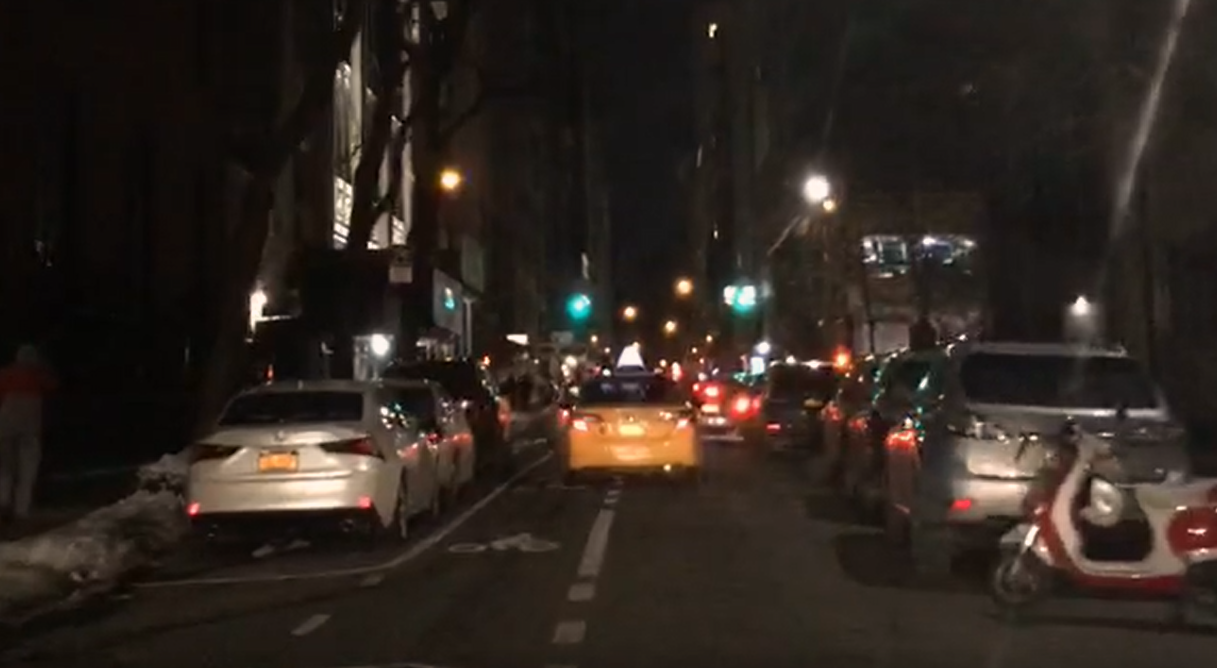} \\
            GT & because traffic is moving now \\
            T=0 & because the light is green and traffic is moving \\
            T=0.3 & as the light turns green and traffic is moving \\ %
            T=0.3 & because the light is green and traffic is moving \\ %
            T=0.3 & because traffic is moving forward \\ %
            T=0.3 & because the light turns green \\ %
            T=0.3 & because the light turned green and traffic is moving \\ %
            \bottomrule
        \end{tabular}}
        \caption{\textbf{Various samples of generated explanations}. GT stands for the the ground-truth (human gold label). Other lines are justifications generated by the model BEEF, obtained with with various decoding temperature T: T=0 corresponds to the greedy decoding and the lines with T=0.3 correspond to random decoding with a temperature of 0.3.
        \updated{A higher temperature implies an increased diversity in the generated justifications, both syntactically and in the completeness level, with some sentences mentioning the light and/or the moving traffic. However, pushing this parameter too high may also lead to an undesired semantic drift.} 
        Credits to \citep{beef}.}
        \label{tab:sec:6:beef_temp}
    \end{table}

    Interestingly, \citet{beef} draw a parallel between VQA \citep{vqa,vqa_ijcv,ask_your_neurons} and the task of explaining decisions of a self-driving system \updatedlast{with natural language}: similarly to the way the question is combined with visual features in VQA, in their work, decisions of the self-driving system are combined with perceptual features encoding the scene. For the VQA task, the result is the answer to the question and, in the case of the driving explanations, the result is the justification why the self-driving model produced its decision. 
    More generally, we believe that recent VQA literature can inspire more explainable driving works.
    In particular, there is a strong trend to make VQA models more interpretable \citep{vqa_e,explaining_vqa_predictions,study_multimodal_interactive_vqa}, to unveil learned biases \citep{dont_just_assume,overcoming_language_priors,rubi}, and to foster reasoning mechanisms \citep{clevr,learning_to_reason,murel}.
    Lastly, towards the long-term goal of having human-machine dialogs and more interactive explanations, the VQA literature can also be a source of inspiration \citep{study_multimodal_interactive_vqa}.
    
    We remark that driving datasets that are designed for explainability purposes have poor quality on the automated driving side. For instance, they include only one camera, the sensor calibration is often missing, \textit{etc}.
    We argue that better explainability datasets should be proposed, by building on high-quality driving datasets, such as nuScenes \citep{CaesarBLVLXKPBB20}. Regarding the lack of high-quality driving datasets containing explanations, another research direction lies in transfer learning for explanation: the idea would be to separately learn to drive on big driving datasets and to explain on more limited explanation datasets. The transfer between the two domains would be done by fine-tuning, by using multi-task objectives, or by leveraging recent transfer learning works.
    
    \subsection{Limits of mimicking natural language explanations.}
    \label{sec:6:limits}

    Using annotations of explanations to supervise the training of a neural network seems natural and effective. 
    Yet, this practice has some strong assumptions and the generated explanations may be limited in their faithfulness.
    From a data point-of-view, as was noted in \citep{textual_explanations}, acquiring the annotations for explanations can be quite difficult: ground-truth explanations are often post-hoc rationales generated by an external observer of the scene and not by the person who took the action. 
    Beyond this, explanation annotations correspond to the reasons why a \emph{person} made an action. Using these annotations to explain the behavior of a \emph{machine learning model} is an extrapolation that should be made carefully. Indeed, applying some type of behavior cloning method on explanations assumes that the reasons behind the model decision must be the same as the one of the human performing the action.
    This assumption prevents the model to discover new cues on which it can ground its decision.
    For example, in medical diagnosis, it has been found that machine learning models can discover new visual features and biomarkers, which are linked to the diagnosis through a causal link unknown to medical experts \citep{differences_human_machine_perception}.
    In the context of driving, however, it seems satisfactory to make models rely on the same cues human drivers would use. 
    
    Beyond the aforementioned problems, evaluating natural language explanations constitutes a challenge per se.
    Most approaches \citep{textual_explanations,generating_visual_explanations,esnli,explain_yourself} evaluate generated natural language explanations based on human ratings or by comparing them to ground-truth explanation of humans (using automated metrics like BLEU \citep{bleu}, METEOR \citep{meteor}, or CIDEr \citep{cider} scores).
    As argued by \citet{leakage_adjusted_simulatability,GilpinBYBSK18}, the evaluation of natural language explanations is delicate and automated metric and human evaluations are not satisfying as they cannot guarantee that the explanation is faithful to the model's decision-making process.
    These metrics rather evaluate the plausibility of the explanation regarding human evaluations \citep{aligning_faithful_interpretations}.
    
    Overall, this evaluation protocol encourages explanations that match human expectation and it is prone to produce \emph{persuasive explanations} \citep{promise_and_peril_human_eval,GilpinBYBSK18}, \ie explanations that satisfy the human users regardless of their faithfulness to the model processing.
    Similarly to what is observed in \citep{sanity_check_saliency} with saliency maps, the human observer is at risk of confirmation bias when looking at outputs of natural language explainers.
    Potential solutions to tackle the problem of persuasive explanations can be inspired by recent works in NLP. Indeed, in this field, several works have recently advocated for evaluating the \emph{faithfulness} of explanations rather than their \emph{plausibility} \citep{towards_faithfully_interpretable_nlp,evaluating_saliency_methods_nl}.
    For example, \citet{leakage_adjusted_simulatability} propose the leakage-adjusted simulatability (LAS) metric, which is based on the idea that the explanation should be helpful to predict the model's output without leaking direct information about the output.

\section{Conclusion}
\label{sec:conclusion}

    In this survey, we presented the challenges of explainability raised by the development of modern, deep-learning-based self-driving models.
    In particular, we argued that the need for explainability is multi-factorial, and it depends on the person needing explanations, on the person's expertise level, as well as on the available time to analyze the explanation.
    We gave a quick overview of recent approaches to build and train modern self-driving systems and we specifically detailed why these systems are not explainable \textit{per se}.
    First, many shortcomings come from our restricted knowledge on deep learning generalization, and the black-box nature of learned models. Those aspects do not spare self-driving models.  
    Moreover, as being very heterogeneous systems that must simultaneously perform tasks of very different natures, the willingness to disentangle implicit sub-tasks appears natural.
    
    As an answer to such problems, many explanation methods have been proposed, and we organized them into two categories.
    First, \emph{post-hoc} methods which apply on a trained driving model to locally or globally explain and interpret its behavior.
    These methods have the advantage of not compromising driving performances since the explanation models are applied afterward; moreover, these methods are usually architecture-agnostic to some extent, in the sense that they can transfer from a network to another one. However, even if these techniques are able to exhibit spurious correlations learned by the driving model, they are not meant to have an impact on the model itself.
    On the other hand, directly \emph{designing} interpretable self-driving models can provide better control on the quality of explanations at the expense of a potential risk to degrade driving performances.
    Explainability is contained in the neural network architecture itself and is generally not transferable to other architectures

    Evaluating explanations \updated{is a huge challenge.
    Automated evaluation methods usually depend on the explainability scheme and the type of data, and there is a lack of unified evaluation methods.
    Concerning crow-sourced evaluations, for example to evaluate natural language explanations with a human rating, is not satisfying as it does not scale well for future methods and as it can lead to persuasive explanations, especially if the main objective is to increase users' trust.
    } 
    In particular, this is a serious pitfall for approaches that learn to mimic human explanations (\eg imitation learning for explanations) such as models in \citep{textual_explanations,generating_visual_explanations,multimodal_explanations}, but also for post-hoc saliency methods \citep{sanity_check_saliency}.
    A solution to this issue could be to measure and quantify the \emph{uncertainty of explanations}, \ie answering the question ``\textit{how much can we trust explanations?''}. Related to this topic is the recent work of \citet{confidnet}, which learns the confidence of predictions made by a neural network with an auxiliary model called ConfidNet, or the work of \citet{uncertainty_explanation} which applies explanation methods to Bayesian neural networks instead of classical deep networks, thus providing built-in modeling of uncertainties for explanations.
    \updated{
    Besides, evaluating \emph{complete} explanations of a driving decision is especially challenging. Indeed, more complete explanations may involve \emph{compositional} explanations, \ie that combine several atomic logical facts. Such explanations are particularly hard to evaluate, as argued by \citet{multi_hop_explanation_evaluation}, because they cannot be satisfyingly evaluated against gold-labels as the number of combinations explodes with the number of facts.
    }
    Overall, finding ways to evaluate explanations with respect to key concepts such as human-interpretability, completeness level, or faithfulness to the model's processing is essential to design better explanation methods in the future.

    Writing up this survey, we observe that many X-AI approaches have not been used --- or in a very limited way --- to make neural driving models more interpretable.
    This is the case for example for local approximation methods, for counterfactual interventions, or model translation methods. 
    Throughout the survey, we hypothesized the underlying reasons that make it difficult to apply off-the-shelf X-AI methods for the autonomous driving literature. 
    One of the main hurdles lies in the type of input space at hand, its very high dimensionality, and the rich semantics contained in a visual modality (video, 3D point clouds).
    Indeed, many X-AI methods have been developed assuming either the interpretability of each of the input dimensions or a limited number of input dimensions.
    Because of the type of the input space for self-driving models, many X-AI methods do not trivially transpose to make self-driving models more interpretable.  
    For example, one will obtain meaningless adversarial perturbations if naively generating counterfactual explanations on driving videos and we thereby observe a huge gap between the profuse literature for generating counterfactual examples for low-dimensional inputs and the scarce literature on counterfactual explanations for high-dimensional data (images and videos).
    As another example, it seems impractical to design a sampling function in the video space to locally explore around a particular driving video and learn a local approximation of the self-driving model with methods presented in \autoref{sec:4:local:saliency}.
    We believe that ways to bridge this gap, detailed in \autoref{sec:5:input:semantic}, include making raw input spaces more controllable and manipulable, and designing richer input semantic spaces that have human-interpretable meaning. 
    
    Despite their differences, all the methods reviewed in this survey share the objective of exposing the causes behind model decisions.
    Yet, only very few works directly borrow tools and concepts from the field of causal modeling \citep{pearl_causality}. 
    Taken apart methods that attempt to formulate counterfactual explanations, applications of causal inference methods to explain self-driving models are rare.
    As discussed in \autoref{sec:4:local:counterfactual}, inferring the causal structure in driving data has strong implications in explainability. It is also a very promising way towards more robust neural driving models. 
    As was stated in \citep{causal_confusion}, a driving policy must identify and rely solely on true causes of expert decisions if we want it to be robust to distributional shift between training and deployment situations. 
    Building neural driving models that take the right decisions for the right identified reasons would yield inherently robust, explainable, and faithful systems.

    \section{Declarations}
    This survey was funded by Valeo and no other funding was received to assist with the preparation of this manuscript.
    The authors have no relevant financial or non-financial interests to disclose.

\bibliographystyle{spbasic}      %
\bibliography{biblio}   %

\end{document}